\documentclass[]{x2lab}

\usepackage{microtype}
\usepackage{graphicx}
\usepackage{float}
\usepackage{subcaption}
\usepackage{tikz}
\usetikzlibrary{arrows.meta,positioning,calc,decorations.pathreplacing,patterns}
\usepackage{csquotes}
\usepackage{afterpage}
\usepackage{placeins}

\usepackage{amsmath}
\usepackage{amssymb}
\usepackage{mathtools}
\usepackage{amsthm}
\usepackage{xspace}
\usepackage{colortbl}
\usepackage{multirow}
\usepackage{enumitem}
\usepackage{wrapfig}
\usepackage{nicefrac}
\usepackage{makecell}

\usepackage{wrapfig}
\usepackage{xcolor}
\usepackage{tikz}
\usetikzlibrary{arrows.meta, decorations.pathreplacing}

\usepackage{pifont}

\definecolor{fbApp}{HTML}{c8e7fa}
\definecolor{fbPurple3}{HTML}{f0ebf5}

\definecolor{citecolor}{HTML}{0071BC}
\definecolor{linkcolor}{HTML}{ED1C24}
\definecolor{efblue}{RGB}{0, 102, 204}   

\providecommand{\ours}{\textsc{WALL-SS}\xspace}

\makeatletter
\renewcommand\paragraph{\@startsection{paragraph}{4}{\z@}%
  {1.5ex \@plus 1ex \@minus .2ex}%
  {-1em}%
  {\normalfont\normalsize\bfseries}}
\makeatother

\title{WALL-SS: Scaling Long-horizon World Models via Next-Scale Autoregression}
\author{X Square Robot Team}

\abstract{
Generative world models provide robots with predictive models of how the world evolves under interaction, with growing potential for simulation, planning, policy evaluation, and robot learning. Moving beyond clip-level future prediction, however, calls for a unified generative formulation that explicitly relates actions to their consequences, supports predictions over flexible horizons, maintains world state during continuous interaction, and provides a probabilistic interface for reward-driven optimization. We introduce \textbf{\ours{}}, a world model that generates visual futures through \textbf{S}cale-wise autoregressive \textbf{S}caling, enabling action-controllable and long-horizon robotic simulation. \ours{} represents embodied trajectories as causal sequences of temporally interleaved observations and actions, making action-dependent state transitions explicit while naturally supporting variable-length generation, streaming extension through reusable causal states, and direct optimization through sequence probabilities. To make this formulation effective over long horizons, we generate each future observation in a coarse-to-fine manner and develop three complementary components within the same hierarchy. Action-conditioned next-scale prediction introduces scale-aligned action representations throughout generation to strengthen action--future coupling and faithfully simulate diverse prescribed behaviors, including suboptimal and failed outcomes. Scale-compressed long-horizon memory preserves recent interactions at fine resolution while progressively compressing distant observations and their corresponding action histories under a bounded memory budget, with scale-wise dream forcing improving robustness to self-generated context. Finally, on-policy alignment of autoregressive visual dynamics treats next-scale generation as a stochastic policy and optimizes fresh rollouts using action-following and long-term consistency rewards while preserving the pretrained visual distribution. Experiments show that \ours{} improves action following and trajectory accuracy, supports coherent minute-long streaming rollout under bounded memory, and consistently benefits from on-policy alignment in reducing action drift and long-horizon inconsistency. Moreover, closed-loop deployment of the same external robot policy in \ours{} and the physical world yields calibrated task outcomes and consistent policy-checkpoint rankings, demonstrating that the learned world model preserves action-dependent dynamics relevant to downstream robot policy evaluation.

}
  
\date{\today}

\metadata[Code]{\url{https://github.com/X-Square-Robot/wall-ss}}
\metadata[Project Page]{\url{http://x2robot.com/pages/ss}}

\begin{document}
\maketitle

\section{Introduction}
\label{sec:introduction}

\begin{figure*}[t]
\centering
\includegraphics[width=\textwidth]{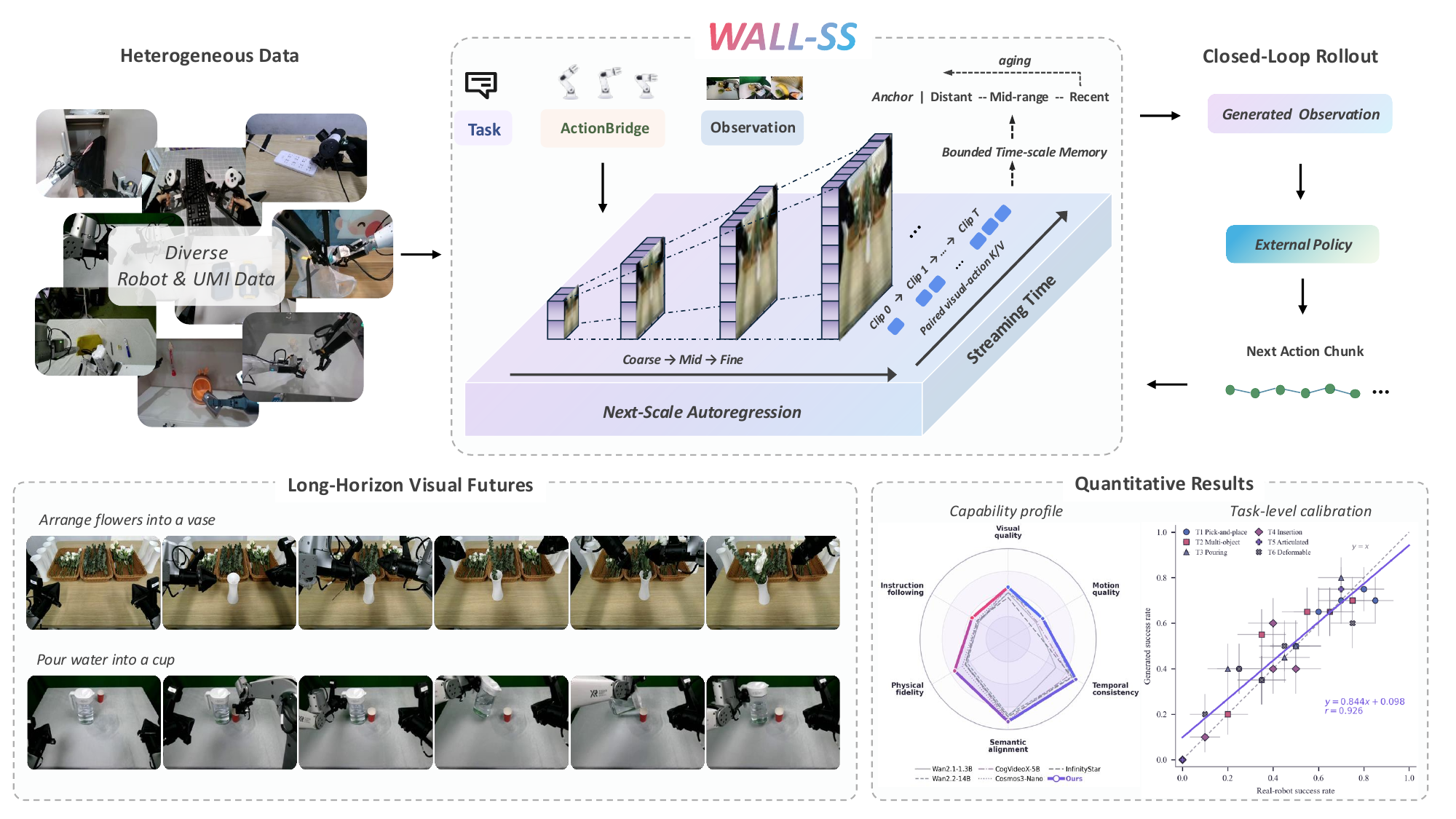}
\caption{\textbf{Overview of \ours{}.}
\ours{} learns action-grounded visual dynamics from heterogeneous robot and
UMI demonstrations. Given a task instruction, an initial multi-view
observation, and prescribed robot actions, the model generates visual futures
through coarse-to-fine next-scale autoregression and propagates state over
streaming time using bounded time--scale memory. Generated observations can be
recursively consumed by an external robot policy for closed-loop rollout.
Bottom panels show representative long-horizon visual futures and quantitative
evaluations of generation capability and real-robot calibration.}
\label{fig:overview}
\end{figure*}

Generative world models aim to capture how embodied experience evolves under interaction. Their potential in robotics extends beyond serving as visual simulators: by relating what an agent observes, intends, and does over time, they can provide reusable predictive structure for planning, data generation, policy evaluation, policy optimization, and robot learning~\cite{black2024pi_0,kim2024openvla,bjorck2025gr00t,kang2026xtokenizer}. Predicted future observations make the consequences of candidate actions visually inspectable, while the learned dynamics provide a complementary interface to vision--language--action models for reasoning about how actions transform the world. Recent video world models have made impressive progress in visual realism, temporal coherence, and action controllability, establishing action-conditioned video generation as a promising foundation for embodied intelligence. As world prediction becomes increasingly connected with action generation and planning, faithfully modeling how actions affect future world states becomes increasingly important. These advances motivate a broader question: how can high-quality action-conditioned prediction be developed into a general generative model of embodied trajectories? Such a model should explicitly organize actions and their consequences within a shared causal history, accommodate predictions of different durations and granularities, incrementally maintain world state during continuous interaction, and expose a probabilistic interface for reward-driven optimization. Many existing systems are primarily organized around clip-level future prediction, with these broader capabilities supported through separate, task-specific interfaces. The remaining challenge is therefore not action conditioning alone, but finding a unified modeling formulation that connects perception, action, memory, and optimization within a single evolving generative process.

\begin{figure*}[t]
    \centering
    \includegraphics[width=0.75\textwidth]{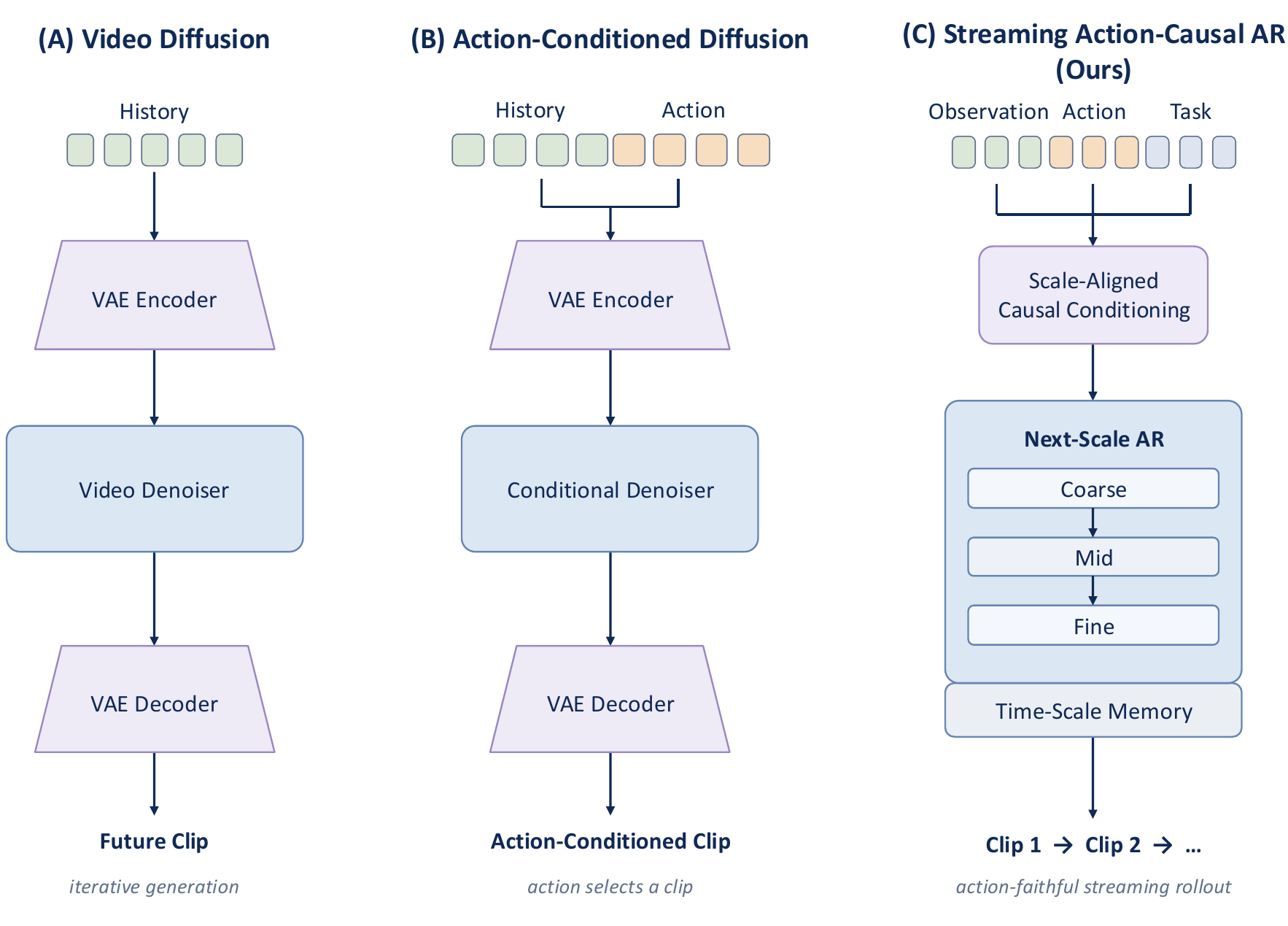}
    \caption{\textbf{Comparison of generative world-model formulations.}
    (a) Conventional video diffusion predicts a future clip from visual
    history through iterative denoising.
    (b) Action-conditioned diffusion additionally conditions generation on
    robot actions, but is typically organized as a clip-level prediction
    process.
    (c) In contrast, \ours{} causally aligns observations, actions, and task
    context, generates visual futures through coarse-to-fine next-scale
    autoregression, and propagates bounded time--scale memory across clips for
    action-faithful streaming rollout.}
    \label{fig:architecture-comparison}
\end{figure*}

Fig.~\ref{fig:architecture-comparison} highlights the structural difference between conventional video diffusion, action-conditioned diffusion, and our streaming action-causal formulation. Although action-conditioned diffusion introduces robot actions into video generation, prediction is still commonly organized around individual future clips. In contrast, we formulate robotic world evolution as a causal sequence of task conditions and temporally interleaved observations and actions. Preserving this temporal ordering makes the correspondence between actions and state transitions explicit in the model factorization: each future observation is generated from the preceding world history together with the intervening actions. This causal formulation provides four key advantages. \textbf{(1) Unified representation.} Vision, language, and action are modeled within a shared representation space and causal architecture, allowing actions and their accumulated consequences to be represented within the same evolving context. \textbf{(2) Variable-length generation.} Future representations can be generated at varying durations and levels of granularity without imposing a fixed prediction horizon or output structure. \textbf{(3) Streaming extension.} Incremental generation and reusable causal states naturally support continuously running systems with online memory maintenance and interaction. \textbf{(4) Direct reward optimization.} Explicit sequence probabilities and token-level log-probabilities provide a natural interface for reinforcement learning and other reward-based objectives. The same causal representation can further support both action-conditioned simulation, where future controls are prescribed, and task-directed imagination, where future actions can be generated from the current state and instruction. These capabilities, however, come with a fundamental autoregressive challenge: every generated observation becomes part of the context for subsequent prediction, allowing small errors to accumulate over time. Such errors lead not only to visual drift, but also to confusion over persistent world states, object identities, interaction outcomes, and the action histories responsible for them. Once this causal history becomes inconsistent, subsequent predictions are generated from an incorrect world state, progressively undermining long-horizon coherence and action fidelity. Autoregressive world modeling is therefore a particularly promising yet demanding direction: it provides a general foundation for unified, flexible, streaming, and reward-optimizable embodied modeling, while making error accumulation and coherent long-term memory central technical challenges.

Building on the analysis above, motivated by next-scale prediction, we propose \ours{}, a next-scale autoregressive world model for action-controllable and long-horizon robotic simulation (Fig.~\ref{fig:overview}). We factorize each future into a sequence of coarse-to-fine predictions. This next-scale autoregression provides a structured way to model world evolution at multiple spatiotemporal granularities. At the same time, it retains the unified, variable-length, streaming, and reward-optimizable properties of autoregressive modeling. Building on this formulation, \ours{} integrates action-conditioned next-scale prediction, scale-compressed long-horizon memory, and reward-based autoregressive post-training to strengthen action--future coupling, reduce long-horizon drift, and improve the physical consistency of generated futures. We now detail how these three components are instantiated within a unified next-scale framework.

We first address action--future coupling through \textbf{action-conditioned next-scale prediction}, which introduces scale-aligned action representations throughout coarse-to-fine generation, strengthening the coupling between robot controls and predicted state transitions. This enables the model to simulate scene evolution under diverse candidate action sequences, including suboptimal and failed behaviors. Importantly, such action grounding requires the generated future to follow the prescribed controls even when they deviate from the task or scene prior, allowing the model to represent counterfactual action-dependent outcomes rather than only likely successful continuations. It therefore improves generation flexibility and encourages the world model to represent a broader range of possible dynamics rather than only high-probability successful futures.

Second, \textbf{scale-compressed long-horizon memory} connects short-term and long-term memory with the same coarse-to-fine structure used in next-scale prediction, creating a unified and self-consistent hierarchy between how the model predicts the future and how it remembers the past. Recent history is kept at fine detail to preserve motion and interaction information, while older history is gradually compressed into coarser summaries of stable world states. Visual memories and their corresponding action histories are compressed in the same way, so that each remembered state remains linked to the actions that produced it. This design makes it clear what information should be retained at different temporal ranges and enables stable long-horizon rollout under a limited memory budget. To ensure that the model can reliably use such memories during autoregressive inference, we further introduce scale-wise dream forcing, which trains the model on corrupted and self-generated historical contexts to reduce exposure bias and accumulated rollout errors.

Third, \textbf{on-policy alignment of autoregressive visual dynamics} treats next-scale visual-token generation as a stochastic policy. Under fixed robot-action conditions, fresh rollouts sampled from the current visual policy are optimized with action-following and long-term consistency rewards, while autoregressive replay and reference-model regularization preserve the pretrained visual distribution. To evaluate the resulting dynamics at the system level, we deploy the external robot policy in closed loop in both \ours{} and the physical world, and compare their long-horizon task outcomes.

Experiments show that \ours{} improves action following and trajectory accuracy, supports minute-long streaming rollout under a bounded memory budget, and benefits consistently from on-policy alignment in reducing action drift and long-horizon inconsistency. We further observe agreement between simulated and physical closed-loop execution, including calibrated task outcomes and consistent policy-checkpoint rankings, providing direct evidence that the learned world model preserves action-dependent dynamics relevant to downstream policy evaluation.

Together, these designs establish \ours{} as an action-controllable, long-horizon, and reward-aligned autoregressive world model. Our main contributions are:

\begin{itemize}[leftmargin=*,topsep=0.35em,itemsep=0.35em]

\item \textbf{Next-scale autoregression for action-controllable world modeling.}
To the best of our knowledge, we present the first robotic world model based on next-scale autoregression. We introduce scale-aligned action conditioning to guide coarse-to-fine world generation, supporting diverse futures that include both successful and failed behaviors.

\item \textbf{Scale-compressed memory for long-horizon world modeling.}
We align the temporal hierarchy of short- and long-term memory with the representation hierarchy of next-scale prediction. Fine recent history, coarse distant state, and temporally aligned action history enable interpretable long-horizon reasoning under a bounded inference budget, while scale-wise dream forcing improves robustness to self-generated context.

\item \textbf{On-policy alignment of autoregressive visual dynamics.}
We formulate next-scale visual generation as a stochastic policy and optimize fresh rollouts under fixed robot-action conditions using action-following and long-term consistency rewards. Autoregressive replay and reference-model regularization preserve the pretrained visual distribution during reward optimization.

\end{itemize}

\section{Related Work}
\label{sec:related-work}

\subsection{Generative World Models for Physical Interaction}

World models predict the consequences of interaction. PlaNet, Dreamer, and DreamerV3~\cite{hafner2019learning,hafner2019dream,hafner2023mastering} learn latent dynamics for planning or policy learning, while V-JEPA~2~\cite{assran2025v} learns action-free predictive representations and V-JEPA~2-AC adds latent action-conditioned prediction for goal-image planning. Generative embodied models provide a complementary, inspectable interface by rendering future geometry, motion, contact, and outcomes.
Control enters these models at different stages. IRASim~\cite{zhu2024irasim} conditions video generation on frame-aligned robot trajectories, whereas UniSim~\cite{yang2024unisim} simulates visual outcomes under high-level instructions or low-level controls. LaDi-WM~\cite{huang2025ladi} instead predicts future VFM-aligned latent states to guide a diffusion policy; AdaWorld~\cite{gao2025adaworld} learns transferable latent actions from unlabeled video. Together, they make future dynamics controllable or actionable, reducing ambiguity in contact-rich transitions.

The inverse direction translates visual futures into robot commands. UniPi and RoboDreamer pair generated video plans with inverse-dynamics policies, while dense-\hspace{0pt}correspondence planning derives commands from optical flow and depth between predicted frames~\cite{du2023learning,zhou2024robodreamer,ko2024learning}. A growing line further unifies prediction and control. Unified World Models and Motus support forward, inverse, policy, and joint video--action modes; MotuBrain and Causal World Modeling couple video representations with action prediction; Cosmos Policy generates actions, future states, and values with a video-model backbone~\cite{zhu2025unified,bi2025motus,team2026motubrain,li2026causal,kim2026cosmospolicy}; and Cosmos~3~\cite{nvidia2026cosmos3} and WALL-WM~\cite{li2026wallwm} provide multimodal or event-grounded world--action formulations. These directions complement action-conditioned forward prediction, where controls are inputs and future video is the prediction target.


\subsection{Autoregressive Video Generation}

Autoregressive video models factorize trajectories into causal predictions, supporting heterogeneous conditions, variable-length outputs, reusable states, and explicit likelihoods. VideoGPT autoregressively models discrete spatiotemporal latents; Phenaki instead uses a temporally causal tokenizer to encode and decode variable-length video, together with a bidirectional masked Transformer to generate video tokens~\cite{yan2021videogpt,villegas2022phenaki}. VideoPoet and Emu3 extend autoregression across modalities~\cite{kondratyuk2024videopoet,wang2024emu3}, and Cosmos~\cite{nvidia2025cosmosworldfoundationmodel} supports action-conditioned recursive rollout. To reduce token-wise decoding, NOVA predicts frames or frame sets without vector quantization and MAGI-1 operates over temporal chunks, improving parallelism while retaining incremental generation and cache reuse~\cite{deng2024nova,sand2025magi}.

Next-scale models instead predict structured token maps from coarse to fine. VAR~\cite{tian2024var} parallelizes variables within each scale; FVAR~\cite{li2025fvar} uses progressive optical refocusing and high-frequency residual distillation to suppress aliasing without changing the inference architecture; and Infinity~\cite{han2025infinity} adds bitwise multi-scale residual quantization and self-correction so later scales can compensate for earlier errors. InfinityStar~\cite{liu2025infinitystar} extends the hierarchy to video through an initial-frame image pyramid and fixed-duration clip pyramids, refining each clip across scales and generating clips autoregressively over time. This introduces causal axes over both clip and scale.


Generative probabilities also enable alignment. DDPO and Diffusion-DPO optimize diffusion generators with reward or preference feedback, while PhysHPO targets physical and semantic video properties~\cite{black2024ddpo,wallace2024diffusiondpo,chen2025physhpo}. RLVR-World, RLIR, and EVA target verifiable decoded predictions, control--video agreement, or executable plans~\cite{wu2025rlvrworld,ye2025rlir,wang2026eva}; WorldGym and WMPO use learned dynamics for policy evaluation or optimization~\cite{quevedo2025worldgym,zhu2025wmpo}. Our method directly optimizes the visual-token distribution at selected action-sensitive scales using action, transition, and task-progress feedback, while likelihood replay and reference regularization preserve fine appearance.

\subsection{Long-Horizon World Modeling}


Long-video models balance recent and persistent context in different ways: StreamingT2V~\cite{henschel2025streamingt2v} retains an appearance anchor; FramePack~\cite{zhang2025framepack} compresses older frames; LongLive~\cite{yang2025longlive} combines local windows with frame sinks; and VideoSSM~\cite{yu2025videossm} adds recurrent state-space memory. These designs preserve recent motion at higher resolution than long-term appearance.

Structured memory methods instead use geometry-grounded maps and episodic frames in Long-term Spatial Memory~\cite{wu2025spatialmemory}; filter history in StableWorld~\cite{yang2026stableworld}; organize bounded anchor, history, and window caches in Echo-Forcing~\cite{wu2026echoforcing}; or consolidate evicted key--value states in Echo-Infinity~\cite{bian2026echoinfinity}. These mechanisms explicitly select or consolidate history rather than retaining an undifferentiated context.

A separate line addresses exposure bias: teacher-forced models train on clean history but infer from their own predictions~\cite{bengio2015scheduled}. Diffusion Forcing trains with independently noised ground-truth tokens, covering heterogeneous context-noise patterns without directly exposing the model to its own rollout errors~\cite{chen2024diffusionforcing}. Self-Forcing, Rolling Forcing, and Resampling Forcing more directly align training with autoregressive inference through self-generated rollouts, rolling windows trained on self-generated histories, or self-resampled degraded histories, respectively~\cite{huang2025selfforcing,liu2025rollingforcing,guo2025resamplingforcing}, reducing exposure bias and long-horizon error accumulation.



\section{Architecture Design: Next-Scale Autoregressive World Modeling}
\label{sec:method}
\subsection{Overview}
\label{sec:method-overview}
\begin{figure*}[t]
    \centering
    \includegraphics[width=\textwidth]{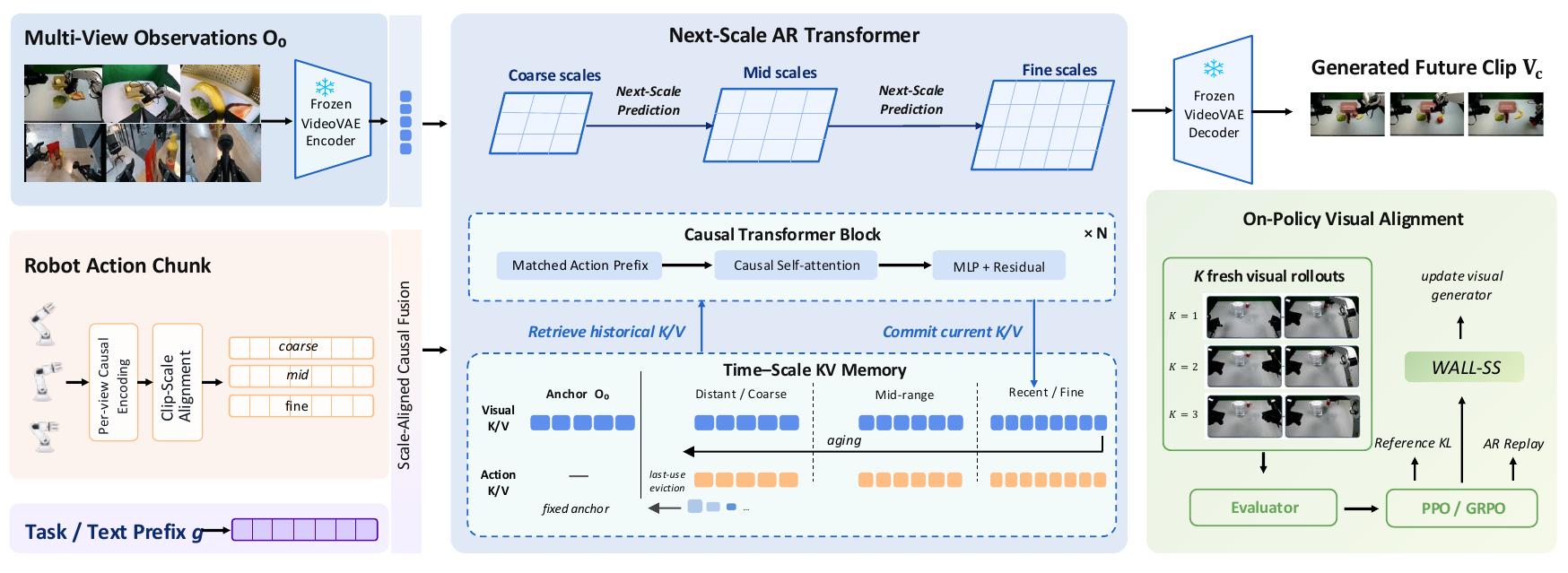}
    \caption{\textbf{Overall framework of \ours{}.}
    \ours{} unifies action-conditioned next-scale visual prediction,
    streaming long-horizon state propagation, and on-policy alignment of
    autoregressive visual dynamics.}
    \label{fig:framework}
\end{figure*}

As summarized in Fig.~\ref{fig:framework}, \ours{} instantiates
action-grounded next-scale autoregressive world modeling with a visual
generator initialized from InfinityStar.  Given an
initial multi-view observation, a task instruction, and temporally aligned
controls, the model advances physical time and constructs each
synchronized future from coarse latent structure to fine residual detail.  A
deterministic grounding interface projects controls onto the corresponding
camera timelines, while scale-aligned causal masks expose only the actions
associated with the current visual interval.
To support recursive rollout, \ours{} reuses causal states formed within the
same next-scale hierarchy as streaming memory, retaining recent observations
at finer scales and distant history at coarser scales while preserving the
initial observation as a persistent anchor. 
Since the visual generator defines explicit autoregressive probabilities, it
can further be aligned on its own rollouts using feedback on action following
and long-horizon consistency.

The remainder of this section develops these components in sequence.  First,
Sec.~\ref{sec:action-grounded-next-scale} introduces the next-scale visual
representation, action--video temporal grounding, and scale-aligned causal
conditioning used to learn action-dependent transitions.  Second,
Sec.~\ref{sec:scale-compressed-memory} converts the same hierarchy into bounded
time--scale memory and trains it to remain stable under recursively generated
history.  Third, Sec.~\ref{sec:world-action-policy} adds the auxiliary action
expert, which predicts executable controls from committed world states and
shares the grounding interface used by external actions.  Finally,
Sec.~\ref{sec:fidelity-visual-alignment} aligns the visual generator on self-generated trajectories while preserving its
pretrained visual prior.

\subsection{Action-Grounded Next-Scale Dynamics}
\label{sec:action-grounded-next-scale}

Visual priors can generate plausible futures while underusing controls that
are correlated with the scene or instruction.  We therefore inject actions
into the causal decisions that construct each future state, rather than using
them as a single global prompt.

\paragraph{Next-scale visual representation.}
Each scale predicts residual visual codes over the cumulative reconstruction
from preceding scales.  The matched action remains available throughout this
coarse-to-fine refinement, and a residual cross-view adapter exchanges only
causally available synchronized states, allowing all cameras to constrain the
same transition without a camera-wise autoregressive order.

\paragraph{Action--video temporal grounding.}
For each camera $v$, a deterministic renderer $\mathcal R^{(v)}$ projects the
plan-side trajectory, initialized from rollout state $p_0$, into the visual
timeline.  It uses only prescribed controls and calibration, never realized
future telemetry or target video.  A shared causal encoder processes this
stream continuously and slices it by the same clip intervals $\mathcal J_c$
used for RGB, preserving both temporal continuity and clip-level alignment.

The causal action encoder produces a continuous feature stream for each view.
Features aligned with clip interval $\mathcal J_c$ are projected at a retained
set of action scales $\mathcal S_A$ and collected as the synchronized condition
$\mathbf X^A_{c,j}$.  For visual scale $\ell$, $\kappa(\ell)$ selects the
finest retained action scale not exceeding $\ell$.  This read-time selection
does not pool actions over time or mix controls across clips.

\paragraph{Scale-aligned causal conditioning.}
Action tokens are materialized as condition prefixes in the same Transformer
sequence.  A query at $(c,\ell,r,v)$ reads text and only
$(c,\kappa(\ell),v)$ from the block-diagonal action graph; other views,
adjacent clips, and future controls remain masked.  The identity anchor
represents $\mathbf O_0$ alone.  Cross-view exchange acts only on synchronized
causal visual states and leaves this direct action mask unchanged.

\paragraph{Autoregressive learning objective.}
Let $N_{c,\ell,r}$ be the number of valid visual decisions after masking
unavailable camera sites.  Under the same causal order used at inference, we
minimize the balanced negative log-likelihood
\begin{equation}
    \mathcal L_{\mathrm{AR}}
    =-\mathbb E_{\mathcal D}
    \mathbb E_{(c,\ell,r)\sim\mathrm{Bal}}
    \left[
        \frac{1}{N_{c,\ell,r}}
        \log p_\theta\left(
            \mathbf Z_{c,\ell,r}
            \mid\mathbf Z_{c,\prec(\ell,r)},\mathcal M_c,
            \mathbf X^A_{c,\kappa(\ell)},T_g\right)
    \right].
    \label{eq:ar-learning-objective}
\end{equation}
Here $\mathrm{Bal}$ averages clips, scales, and residual repeats uniformly,
while the bundle log-probability sums valid token decisions.  For samples
without reliable actions, the action condition is set to $\varnothing$.

\subsection{Time--Scale Memory for Long-Horizon Rollout}
\label{sec:scale-compressed-memory}

Retaining only the latest generated clip loses persistent state, whereas
retaining all tokens makes memory grow with rollout length.  We instead reuse
the next-scale hierarchy in reverse temporal order: recent clips retain finer
causal states and older clips coarser ones.  These states are captured during
generation, avoiding a separate memory encoder or RGB-space compression.

\paragraph{Time--scale memory records.}
At each retained pre-prediction boundary $s\in\mathcal S_M$, we materialize
the synchronized causal state $\mathbf H^{\mathrm{pre}}_{u,s}$ before the
target at that boundary is observed.  Larger $s$ contains a finer cumulative
prefix.  Each state is paired with the action bank
$\mathsf A_u=\{\mathbf X^A_{u,j}\}_{j\in\mathcal S_A}$ for the same physical
interval:
\begin{equation}
    \begin{aligned}
        \mathbf B_{u,s}
        &=
        \left(
            \mathbf H^{\mathrm{pre}}_{u,s},
            \mathsf A_u
        \right),
        \qquad s\in\mathcal S_M,
        \\
        \mathcal M_c
        &=
        \left\{
            \mathbf H_{\mathrm{anc}}
        \right\}
        \cup
        \left\{
            \mathbf B_{c-\delta,s(\delta)}
            \mid
            \delta\in\mathcal D_c
        \right\},
        \qquad
        \mathcal D_c
        =
        \left\{
            1,\ldots,\min(W_M,c-1)
        \right\},
        \\
        \delta_1<\delta_2
        \Longrightarrow
        s(\delta_1)\geq s(\delta_2).
    \end{aligned}
    \label{eq:time-scale-memory}
\end{equation}
Here $\delta$ is clip age and $W_M$ the fixed retention horizon.  The monotone
schedule selects fine recent and coarse distant records; aging changes only
which precomputed boundary is addressed and never converts KV states online.

\paragraph{Identity-anchored distance-aware retention.}
The persistent anchor $\mathbf H_{\mathrm{anc}}$ is derived only from
$\mathbf O_0$ and stabilizes identity and global layout; it contains no future
action and does not replace the current dynamic state.  Rolling records carry
intermediate state, while each retained visual record remains paired with the
action entry for the same interval.

\paragraph{Bounded memory bank.}
The memory bank includes active records and any precomputed records awaiting
later use.  Each is evicted after its final scheduled read, so a fixed
age--scale schedule gives
\begin{equation}
    \mathcal R_c^{\mathrm{mat}}
    =
    \left\{
        \mathbf H_{\mathrm{anc}}
    \right\}
    \cup
    \left\{
        \mathbf B_{u,s}
        \,\middle|\,
        (u,s)\ \text{is resident before clip }c
    \right\},
    \qquad
    \sup_c
    \operatorname{mem}
    \left(
        \mathcal R_c^{\mathrm{mat}}
    \right)
    \leq
    B_{\mathrm{KV}}.
    \label{eq:bounded-kv}
\end{equation}
Thus persistent history and single-clip working memory remain independent of
the number of completed clips.

\paragraph{Streaming state transition.}
Before clip $c$, the model assembles $\mathcal M_c$, encodes the current
action block, and generates all synchronized views from coarse to fine.
Retained states become persistent only after the complete bundle commits;
the state machine then ages records, changes active cutoffs, and evicts expired
entries before clip $c+1$.

\paragraph{Scale-wise dream forcing.}
Clean teacher forcing does not expose the model to errors stored during
recursive rollout.  We therefore combine age-dependent corruption
$\widetilde{\mathcal M}_c\sim\mathcal C_{\mathrm{age}}(\mathcal M_c)$ with a
detached suffix sampled by a lagged streaming model:
\begin{equation}
    \begin{aligned}
        \widehat{\mathbf Z}_{c:c+d-1}
        &\sim p_{\bar\theta}^{\mathrm{stream}}
        \left(\cdot\mid\mathcal M_c,
        \mathbf X^A_{c:c+d-1},T_g\right),\\
        \mathcal L_{\mathrm{self}}
        &=-\log p_\theta\left(
            \mathbf Z_{c+d}^{\mathrm{gt}}
            \mid\operatorname{sg}\!\left[
            \mathcal U^{(d)}(\mathcal M_c,
            \widehat{\mathbf Z}_{c:c+d-1},
            \mathbf X^A_{c:c+d-1})\right],
            \mathbf X^A_{c+d},T_g\right).
    \end{aligned}
    \label{eq:scale-wise-dream-forcing}
\end{equation}

Let $\mathcal L_{\mathrm{cor}}$ denote the objective with $\widetilde{\mathcal M}_c$ replacing $\mathcal M_c$. The complete objective is
\begin{equation}
    \mathcal L_{\mathrm{long}}
    =
    \mathcal L_{\mathrm{AR}}
    +
    \lambda_{\mathrm{cor}}\mathcal L_{\mathrm{cor}}
    +
    \lambda_{\mathrm{self}}\mathcal L_{\mathrm{self}}.
    \label{eq:long-horizon-objective}
\end{equation}
All branches predict a clean future while varying only the causal history;
dream forcing is therefore supervised exposure-gap training rather than reward
optimization.

\subsection{World Action Policy}
\label{sec:world-action-policy}

The visual model above predicts how an action changes the world.  We complement
this direction with a co-trained action expert.  Before chunk $c$, it reads
$H_c^A=F_\omega^A(\operatorname{Read}_V(\mathcal M_c),E_s(s_c),T_g)$, where
$s_c$ is the current proprioceptive or plan-side state and
$\operatorname{Read}_V$ exposes only committed visual states.  Following
standard flow-matching action experts~\cite{zhai2025igniting,yu2026walloss},
noisy action slots cross-attend to this context and predict the next control
chunk $A_c$.  The mask excludes the current visual target, current action
condition, and all future records.

The executable trajectory $A_c$ remains distinct from the visual condition
$\mathbf X_c^A$.  Predicted and external actions use the same grounding map
$\widehat{\mathbf X}_{c,j}^A=\Gamma_j(\widehat A_c,s_c)$ from
Sec.~\ref{sec:action-grounded-next-scale}; the visual model never reads private
expert states.  Recursive imagination alternates action prediction, visual
generation, and memory commit, whereas physical deployment updates $s_c$ from
measured proprioception.

On paired teacher-forced transitions, both branches read the same pre-action
memory $\mathcal M_c$, and the demonstrated action is both the expert target
and the condition for its recorded visual consequence:
\begin{equation}
    \mathcal L_{\mathrm{cotrain}}
    =
    \mathcal L_{\mathrm{long}}
    +
    \lambda_A\mathcal L_{\mathrm{act}},
    \label{eq:action-video-cotrain}
\end{equation}
where $\mathcal L_{\mathrm{act}}$ follows the cited action-expert objective and
is applied only to paired teacher-forced histories; unpaired histories retain
the visual loss alone.

\subsection{On-Policy Alignment of Visual Dynamics}
\label{sec:fidelity-visual-alignment}

We align the autoregressive visual generator, not the robot controller.
Controls remain fixed conditions while optimization redistributes probability
over the visual futures generated under them.

\paragraph{Visual generator as a policy.}
Let $\xi_n$ contain the discrete prefix, conditions, mask, and memory addresses
at visual-token decision $n$.  A fixed sampling transform defines
$\pi_\theta^V(\cdot\mid\xi_n)=
\mathcal S_\eta[p_\theta(\cdot\mid\xi_n)]$.
For each $x=(\mathbf O_0,\mathbf A_{1:C},g)$, a frozen behavior copy samples
$K\geq2$ trajectories, which are decoded for reward evaluation.  All samples
in a group share the same prescribed controls, so ``on-policy'' refers only to
the visual generation policy.

\paragraph{Dynamics fidelity rewards.}
Two frozen scorers evaluate decoded rollouts.  $R_{\mathrm{act}}$ measures
agreement between each transition and its prescribed action, while
$R_{\mathrm{long}}$ measures state, boundary, and cross-view consistency over
a rolling window.  With $b_c=\max(1,c-W+1)$ and
$\widehat{\mathbf V}^{\,i}_0=\mathbf O_0$,
\begin{equation}
    r_{i,c}
    =
    \lambda_{\mathrm{act}}
    R_{\mathrm{act}}
    \left(
        \widehat{\mathbf V}^{\,i}_{c-1:c},
        A_c
    \right)
    +
    \lambda_{\mathrm{long}}
        R_{\mathrm{long}}
        \left(
            \mathbf O_0,
            \widehat{\mathbf V}^{\,i}_{b_c-1:c},
            \mathbf A_{b_c:c}
        \right).
    \label{eq:dynamics-fidelity-reward}
\end{equation}
Task progress and terminal success are excluded from this reward and reserved
for closed-loop evaluation.

\paragraph{Group-relative policy optimization.}
We aggregate clip rewards as
$G_{i,c}=\sum_{u=c}^{C}\gamma^{u-c}r_{i,u}$ and normalize them across the
$K$ rollouts to obtain
$\widehat A_{i,c}=\operatorname{sg}[
(G_{i,c}-\mu_c)/(\sigma_c+\epsilon)]$.  This supplies a group-relative
baseline without a learned critic.  For token $z_{i,n}$,
\begin{equation}
    \begin{aligned}
        \rho_{i,n}(\theta)
        &=
        \frac{\pi_\theta^V(z_{i,n}\mid\xi_{i,n})}
        {\pi_{\theta_{\mathrm{old}}}^V(z_{i,n}\mid\xi_{i,n})},
        \\
        \mathcal L_{\mathrm{PG}}
        &=-\mathbb E_{\mathrm{bal}}\!\left[
        \min\!\left(
            \rho_{i,n}(\theta)\widehat A_{i,c},
            \operatorname{clip}(\rho_{i,n}(\theta),
            1-\varepsilon,1+\varepsilon)\widehat A_{i,c}
        \right)\right].
    \end{aligned}
    \label{eq:group-relative-visual-pg}
\end{equation}
$\mathbb E_{\mathrm{bal}}$ uses the same clip--scale balancing principle; rewards are detached and gradients pass through
the sampled visual-token log-probabilities.

\paragraph{Prior-preserving regularization.}
We constrain policy drift with a frozen reference generator
$\pi_{\mathrm{ref}}^V$ and real-trajectory autoregressive replay:
\begin{equation}
    \mathcal L_{\mathrm{align}}
    =
    \mathcal L_{\mathrm{PG}}
    +
    \beta
    \mathbb E_{\xi\sim\mathcal D_{\mathrm{roll}}^{\mathrm{bal}}}
    \left[
        D_{\mathrm{KL}}
        \left(
            \pi_\theta^V(\cdot\mid \xi)
            \,\Vert\,
            \pi_{\mathrm{ref}}^V(\cdot\mid \xi)
        \right)
    \right]
    +
    \lambda_{\mathrm{AR}}
    \mathcal L_{\mathrm{AR}}^{\mathrm{real}}.
    \label{eq:visual-alignment-objective}
\end{equation}
The KL term limits local drift, while
$\mathcal L_{\mathrm{AR}}^{\mathrm{real}}$ preserves real-trajectory appearance
and discourages reward exploitation.  Only the visual generator and its
conditioning adapters are updated.

\section{Training Data}
\label{sec:data}

\begin{figure*}[t]
    \centering
    \includegraphics[width=\textwidth]{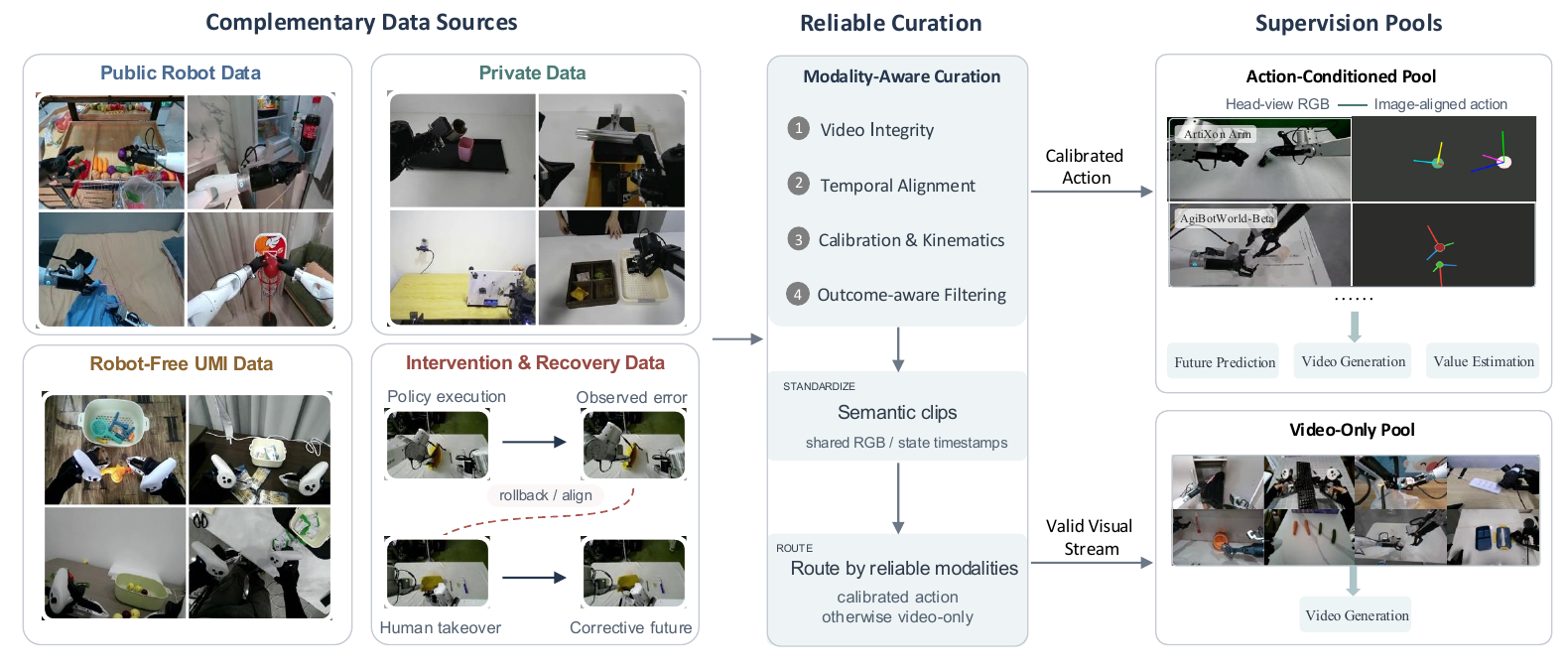}
    \caption{
    \textbf{Overview of the training-data composition and curation pipeline.}
    Public robot data, private robot and non-embodiment UMI recordings,
    and intervention/recovery trajectories are processed through video-integrity,
    temporal-alignment, calibration-and-kinematics, and outcome-aware quality checks,
    and are then standardized into semantic clips at 10 FPS.
    Samples with reliable action geometry are routed to the action-conditioned pool
    $\mathcal{D}_{\mathrm{ac}}$, which supports action-conditioned future prediction,
    video generation, and value estimation.
    Samples with valid visual streams but unavailable or unreliable action geometry
    are retained in the video-only pool $\mathcal{D}_{\mathrm{vid}}$ for video generation.
    }
    \label{fig:data-pipeline}
\end{figure*}

The \ours{} training corpus is designed to support three closely related capabilities: action-conditioned future prediction, text- and first-frame-conditioned video generation, and process-value estimation for policy learning. As shown in Fig.~\ref{fig:data-pipeline}, we organize the corpus into three complementary components: \emph{(i)} public robot data, centered on the full AgiBotWorld-Beta release; \emph{(ii)} privately collected X2-Robot and non-embodiment UMI data; and \emph{(iii)} policy-intervention and failure-recovery trajectories. The public component supplies broad task, scene, and embodiment coverage; the private robot and UMI component reduces the visual, action, and camera-domain gap to our deployment setting; and the intervention component adds off-nominal transitions around contact-rich bottlenecks. This composition follows a general principle also observed in recent embodied world models: broad visual--physical priors and robot-specific controllability are complementary, while both nominal and off-nominal interactions are needed for outcome-faithful rollout~\cite{gigaworld2026roadmap}.

Our data interface follows a \emph{modality-available} principle: each example supervises only the modalities that can be recovered reliably from its source. In particular, action conditioning requires temporally synchronized robot states and controls together with valid camera intrinsics, extrinsics, and kinematic transforms. Samples that lack this calibration are not discarded if their visual stream remains valid. Instead, they remain in the video-generation pool and contribute visual dynamics, temporal structure, and language grounding without an action condition. This separation allows heterogeneous sources to share a common video objective without introducing geometrically misregistered controls.

\subsection{Data Composition}
\label{sec:data-composition}

\paragraph{Public robot data.}
Our principal public source is AgiBotWorld-Beta~\cite{bu2025agibot}, comprising 1,003,672 trajectories collected with over 100 robots across more than 100 real-world scenarios. We use the full release to capture diverse embodiments, objects, scenes, interactions, and task durations. After semantic clip construction and language-annotation validation, the AgiBotWorld-Beta component contains 987,508 captioned clips derived from 165,560 unique source videos. We additionally incorporate real-robot manipulation demonstrations from ManipArena~\cite{sun2026maniparena} to broaden the coverage of reasoning-oriented and contact-rich tasks. Together, these public sources provide a broad embodied prior before adaptation to our internal platforms.

\paragraph{Private data.}
X2-Robot demonstrations are collected in the target observation geometry and contain diverse bimanual manipulation behaviors. When synchronization and calibration checks pass, these trajectories provide robot states, bimanual end-effector motion, gripper state, and multi-view video in the geometry used by the deployed system. They therefore reduce both the embodiment gap and the camera-domain gap between public data and deployment.

\paragraph{Non-embodiment UMI data.}
The non-embodiment subset uses a portable handheld-gripper interface in the spirit of UMI~\cite{chi2024umi}. It expands egocentric hand--object interaction coverage without occupying a robot platform. When the collection rig and cameras are fully calibrated, the recorded motion is converted into the same canonical end-effector representation as the robot data. When this geometric chain is unavailable, the recording remains useful as ordinary embodied video but is not assigned an unreliable action condition.

\paragraph{Policy-intervention and recovery data.}
We maintain policy-intervention and failure-recovery trajectories as a distinct component rather than merging them anonymously into nominal demonstrations. These recordings preserve policy execution, human-controlled corrections, and subsequent interaction under the same collection setup. They expose the model to states that are uncommon under expert teleoperation but frequently reached by imperfect policies, including missed or unstable contacts, misalignment, re-grasping, repeated attempts, and corrective motion. Their temporal structure and supervision roles are described in Sec.~\ref{sec:data-recovery}.

\subsection{Intervention and Failure-Recovery Data}
\label{sec:data-recovery}

Standard robot demonstrations are strongly conditioned on task success: operators typically begin from a valid reset, execute a near-nominal motion, and retain the trajectory when the goal is reached. This distribution is efficient for imitation learning but incomplete for a world model. Near contact, the same high-level intent can produce qualitatively different futures depending on millimeter-scale geometry, friction, compliance, and gripper timing. If almost every gripper-closing action in the training set is followed by a successful lift, a generative model can learn the shortcut that ``grasp'' implies object attachment. It may then move an object into the gripper despite a visible gap, producing an apparently successful but physically implausible, magnet-like grasp.

Intervention and recovery trajectories mitigate this bias by expanding support around the boundary between nominal and unsuccessful interaction. Human corrections are informative because they occur at bottleneck states actually reached by an imperfect policy rather than only at states already well covered by expert demonstrations. Prior intervention-based imitation learning similarly finds that corrections collected at failure states can be more useful than an equivalent amount of non-interventional demonstration data~\cite{mandlekar2021iwr}. Recent embodied world models also retain both successful and failed robot episodes when training or evaluating action-dependent prediction.

Our intervention data contain two complementary collection patterns. In \emph{immediate intervention}, an operator takes control as soon as an error or unsafe trend is detected and continues from the current policy state. This pattern resembles conventional on-policy intervention and captures the states actually visited by the deployed policy. It is particularly useful around fine-manipulation bottlenecks, where small corrections in alignment, contact, wrist motion, or gripper timing can determine the subsequent physical outcome.

In \emph{rollback-and-replay intervention}, the erroneous policy execution is first retained, the system is brought back toward the state immediately before the error, and the operator then executes a corrective control sequence from that rollback point. The resulting record preserves both an erroneous policy branch and a human-corrected branch under closely matched scene, task, and robot conditions. By keeping the high-level context approximately fixed while changing the control sequence, these trajectories provide a hard action-conditioned contrast: the model cannot explain the two futures from scene or task priors alone, but must account for the physical effects of differences in wrist motion, contact geometry, and gripper timing. This supervision is useful for resolving fine-grained, action-conditioned physical effects near contact-sensitive bottlenecks.

Physical rollback does not necessarily reproduce the object pose, contact state, or compliance exactly. We therefore treat the two branches as locally matched intervention examples rather than exact counterfactuals. When paired branches are used for a direct comparison, their consistency is checked from synchronized robot states and visual context around the rollback point. The same structure supports an interpretable qualitative demonstration in which the erroneous policy branch and the human-corrected branch are aligned at the rollback point and shown side by side.

For action-conditioned video prediction, immediate and rollback interventions provide related but distinct supervision. Immediate intervention exposes the model to off-nominal states reached by the policy and to corrective motion from the current state. Rollback-and-replay intervention additionally associates similar high-level context with different control sequences and different physical futures. Together, these trajectories teach three distinctions that nominal demonstrations alone leave underdetermined. First, an intended action is not equivalent to its physical effect: gripper closure without contact must leave the object stationary. Second, failure changes the subsequent dynamics: a slipped, displaced, or partially grasped object creates a new state that cannot be reconstructed from the task instruction alone. Third, correction is itself a structured behavior that may involve retreat, rollback, re-alignment, re-grasping, and renewed contact. Modeling the observed transition discourages a generator from completing a task through a semantically plausible but action-inconsistent visual shortcut.

The same trajectories also provide local temporal ordering for process-value and reward learning. A success-filtered corpus provides little supervision for how value should change when execution degrades and may encourage a nearly monotonic progress estimate after a miss, collision, or drop. Failure and intervention boundaries identify local intervals in which progress has degraded and corrective control begins. When a reliable recovery endpoint or final outcome is available, the subsequent trajectory additionally supervises whether a physically feasible correction restores progress. Policy-rollout data and suboptimal trajectories have similarly been used to refine world-model and value predictions outside expert-only distributions~\cite{kim2026cosmospolicy,wang2026worldvalue}, while physically instantiated failures and recoveries provide richer dense-reward supervision than truncating or relabeling successful demonstrations~\cite{fang2026densereward}. We therefore use local failure and intervention annotations for temporally grounded process supervision, while using recovery-completion and episode-level outcome labels only when they are separately available. A local failure interval is not equated with final episode failure, and a takeover segment is not by itself treated as evidence of successful recovery.

Importantly, retaining off-nominal outcomes does not relax basic quality control. A genuine missed grasp, object drop, collision, retry, rollback, or re-grasp is a valid physical transition and is preserved. Corrupted video, timestamp failure, invalid robot state, or calibration drift is instead a sensing or logging error and is rejected or demoted to video-only supervision. This distinction prevents the cleaning pipeline from removing the off-nominal dynamics that the intervention subset is intended to retain.

\subsection{Data Post-Processing}
\label{sec:data-postprocess}

The raw sources differ in frame rate, camera layout, episode structure, action representation, and calibration availability. We convert them into a common, provenance-preserving episode format through integrity validation, temporal resampling, semantic clip construction, calibration-aware supervision routing, action canonicalization and projection, and offline latent caching. The overall design follows a conservative curation pattern: corrupted and misaligned samples are removed, while useful examples are retained under the subset of modalities they can supervise reliably.

\paragraph{Integrity validation and outcome-aware filtering.}
We first verify frame decodability, resolution and frame-rate metadata, timestamp monotonicity, and the consistency of multi-view streams. Clips with unreadable frames, duplicate-frame collapse, large frame/action length mismatches, missing robot states, non-monotonic control records, or invalid kinematics are rejected. We additionally check for abnormal exposure, severe blur, corrupted compression, camera discontinuities, and truncated episode boundaries. These tests identify acquisition failures rather than task failures: physically valid low-motion intervals such as waiting, holding, and contact stabilization are retained, as are explicit failed attempts, rollbacks, interventions, and recovery motions.

\paragraph{Temporal resampling and clip construction.}
We use semantic subtasks as the basic data unit. Each retained record stores the source video, subtask start and end indices, source frame rate and resolution, language description, outcome and intervention tags when available, and links to action and calibration metadata. Long recordings are segmented at task or subtask boundaries so that one clip expresses a coherent manipulation attempt. Intervention clips retain sufficient pre-error, correction, and rollback context when available to expose the transition that prompted human control. 

\paragraph{Calibration-aware supervision routing.}
Action conditioning is enabled only when the recording contains synchronized actions and a valid geometric calibration chain. Let $m_i^{\mathrm{cal}}\in\{0,1\}$ indicate whether sample $i$ passes the timestamp, camera-intrinsic, camera-extrinsic, and robot-kinematic checks. The curated corpus is routed into
\begin{equation}
    \mathcal{D}_{\mathrm{ac}}
    =
    \{(\mathbf{v}_i,\mathbf{a}_i,\mathbf{l}_i):m_i^{\mathrm{cal}}=1\},
    \qquad
    \mathcal{D}_{\mathrm{vid}}
    =
    \{(\mathbf{v}_i,\mathbf{l}_i):m_i^{\mathrm{cal}}=0\},
    \label{eq:calibration-routing}
\end{equation}
where $\mathbf{v}_i$, $\mathbf{a}_i$, and $\mathbf{l}_i$ denote video, action, and language, respectively. Both subsets train video prediction, but only $\mathcal{D}_{\mathrm{ac}}$ contributes the projected action condition. Missing calibration therefore reduces the available supervision rather than invalidating the visual trajectory. Samples with valid RGB but failed action synchronization or projection are likewise demoted to $\mathcal{D}_{\mathrm{vid}}$ instead of being silently used with a misaligned control.

\paragraph{Action canonicalization and projection.}
For calibrated robot and UMI recordings, heterogeneous actions are first converted into a common end-effector representation. Left- and right-wrist positions and orientations are represented as homogeneous transforms, and gripper opening is normalized to a bounded scalar. Camera extrinsics map each end effector into the head-camera coordinate system, while camera intrinsics are updated to match the aspect-ratio-aware resize and crop applied to RGB frames. For an end-effector keypoint $\mathbf{x}_{e}$, its image coordinate is
\begin{equation}
    \widetilde{\mathbf{u}}
    =
    \mathbf{K}_{h}\,
    \mathbf{T}_{h\leftarrow e}\,
    \widetilde{\mathbf{x}}_{e},
    \qquad
    \mathbf{u}
    =
    \left(
        \widetilde{u}/\widetilde{w},
        \widetilde{v}/\widetilde{w}
    \right),
    \label{eq:action-projection}
\end{equation}
where $\mathbf{K}_{h}$ is the resized head-camera intrinsic matrix and $\mathbf{T}_{h\leftarrow e}$ is the composed transform from the end-effector frame to the head-camera frame. Invalid or behind-camera projections are masked.

We render the projected left- and right-end-effector trajectories as a spatially aligned action video in the main-view image plane. Each frame contains the projected end-effector center and its local orientation axes; color distinguishes the two arms and encodes gripper state, while marker scale reflects camera-relative depth. This representation converts heterogeneous low-dimensional control streams into a video-like condition that shares the RGB stream's timestamps, spatial transforms, and tensor layout. It also makes the calibration requirement explicit: without a valid projection chain, no action video is constructed.

\paragraph{Latent caching and provenance.}
Finally, a frozen video VAE encodes the RGB clip and, when available, its projected action video into temporally aligned latent tensors. Before caching an action-conditioned pair, the preprocessing pipeline checks that RGB and action inputs have identical frame indices and spatial shapes. Cache keys are deterministic functions of the source video, annotated interval, target frame rate, resize/crop configuration, and representation version. Training performs a second cache-existence and tensor-contract check and never falls back to unverified online feature extraction. Each record retains source identity, calibration status, intervention and outcome tags, quality flags, and processing provenance, enabling source-aware sampling and later audits of failure, value, and action-conditioned supervision.

\section{Training and Inference Recipe}
\label{sec:training}

This section describes how the components in Sec.~\ref{sec:method} are optimized
and assembled for rollout.  Starting from an InfinityStar
initialization,
training proceeds through action-grounded autoregressive learning,
rollout-robust long-horizon adaptation, and on-policy alignment of visual
dynamics.  Inference follows the same next-scale order, action alignment, and
streaming memory transition used during training.

\subsection{Autoregressive World-Model Training}
\label{sec:training-autoregressive}

\paragraph{Action-grounded autoregressive learning.}
We train on the curated latent corpus described in Sec.~\ref{sec:data}.  The visual
tokenizer remains frozen, while the autoregressive generator and its action and
view-conditioning modules are optimized jointly.  Calibrated examples provide
the aligned action condition; examples without a reliable action stream retain
their visual likelihood supervision with the action condition absent.
Synchronized camera streams form one multi-view prediction bundle, with
unavailable view sites masked from the likelihood.

Training first uses clean ground-truth contexts and optimizes the next-scale
likelihood.  Each future clip is paired with
its corresponding action chunk, whereas the initial observation is encoded as
an action-free identity anchor.  The shared likelihood trains all synchronized
views and all coarse-to-fine decisions of the same visual future.  The action
expert is co-trained on paired transitions; samples without valid action targets retain only
the visual loss.

\paragraph{Rollout-robust long-horizon adaptation.}
We then continue training on contiguous multi-clip windows with the bounded
time--scale memory enabled.  Clean teacher-forced histories are mixed with
age-aware corrupted histories and detached self-generated suffixes.  The
self-generated context is produced by a lagged model through the same
streaming transition used at inference, while supervision always comes from a
clean future clip as defined by the dream-forcing objective.  This stage
therefore teaches the model to continue from imperfect generated history
without changing the underlying autoregressive objective.

\subsection{On-Policy Alignment of Visual Dynamics}
\label{sec:training-alignment}

\begin{figure*}[t]
    \centering
    \includegraphics[width=\textwidth]{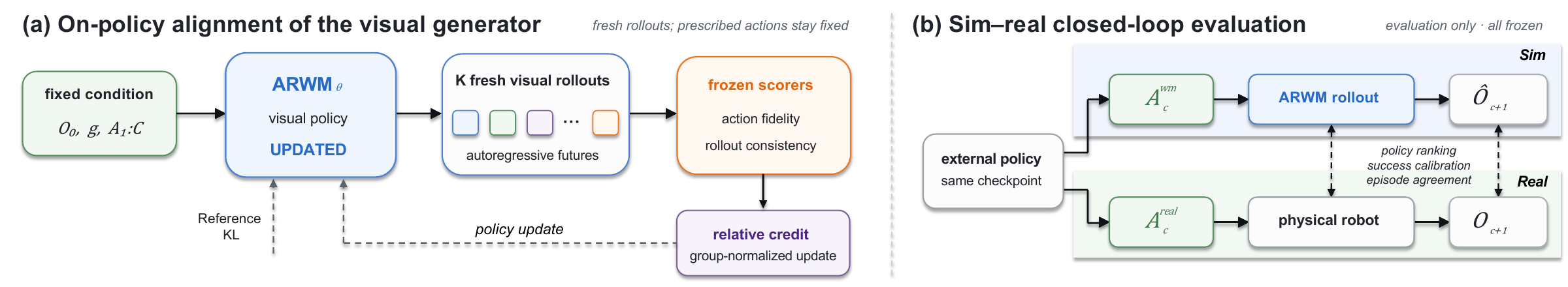}
    \caption{\textbf{On-policy visual alignment and closed-loop evaluation.}
    (a) Under a fixed initial observation, instruction, and prescribed action sequence, a newly frozen behavior model samples fresh long-horizon rollouts.
    (b) At evaluation time, the same frozen external policy interacts
    independently with \ours{} and the physical robot from matched tasks and
    initial states; their outcomes are compared without updating either
    system.}
    \label{fig:rl-closed-loop}
\end{figure*}

Fig.~\ref{fig:rl-closed-loop} summarizes the separation between
reward-based optimization of the autoregressive visual generator and
closed-loop evaluation with a frozen external robot policy.  The former
updates only the visual dynamics under prescribed actions, whereas the latter
tests the resulting world model without optimizing either the controller or
the simulator.

\paragraph{Fidelity scorer training.}
Before policy optimization, we train the action-following and long-horizon
scorers introduced in Sec.~\ref{sec:fidelity-visual-alignment}.  The
former contrasts matched controls with temporally shifted, reversed, or
cross-trajectory actions.  The latter contrasts real contiguous windows with
spliced, drifted, or view-desynchronized sequences.  Both scorers are frozen
after training and provide rewards only for visual-dynamics fidelity.

\paragraph{Visual-policy optimization.}
At each alignment iteration, a just-frozen behavior samples a group of
fresh visual rollouts from the same committed stream state and under the same
action sequence.  Their decoded videos are scored, group-relative returns are
assigned to the sampled visual decisions, and the current model recomputes
their token probabilities to optimize the objective.
All next-scale factors participate with balanced token normalization.
The tokenizer, scorers, reference generator, and any action-producing module
remain fixed, while the autoregressive generator and its conditioning adapters
are updated; rollout batches are refreshed after a bounded number of policy
updates.

\subsection{Inference Recipe}
\label{sec:training-inference}

\paragraph{Action-conditioned streaming.}
Given an initial synchronized observation, a task instruction, and a sequence
of control chunks, the model constructs the identity anchor once and generates
future clips sequentially.  For each clip, it causally encodes the current
action chunk, samples synchronized views in coarse-to-fine next-scale order,
commits the completed bundle, and advances the bounded memory before decoding
the next observation.  The committed states of that clip then condition the
following step.  Independent counterfactual futures can be obtained by forking
the same committed state and supplying different controls or sampling seeds.

\paragraph{Closed-loop rollout.}
For closed-loop validation, a frozen external robot policy consumes the latest
generated synchronized multi-view observation and the task instruction, and
returns the next action chunk.  A lightweight action bridge converts each
policy chunk into the model's control conditioning, resampling it to the
control rate and end-effector representation used during training---after
which \ours{} generates the next synchronized head and both wrist views,
commits them through the streaming state transition, and returns the
observation to the same policy.  After initialization, no ground-truth
observation or action is injected.  The visual generator and external policy
remain fixed throughout evaluation.


\section{Experiments}
\label{sec:arwm-experiments}

\subsection{Embodied Video Generation Evaluation}
\label{sec:embodied-video-generation-eval}
\noindent\textbf{Evaluation protocol.}
To evaluate video-generation capability in embodied settings, we assess \ours{} as an
autoregressive embodied video generator. Given an initial observation and a language
instruction, \ours{} autoregressively rolls out the corresponding future video while
preserving object identity, robot motion, and long-horizon consistency, so that the
generated sequence remains useful for downstream evaluation. Unlike conventional
video-generation benchmarks that focus primarily on perceptual fidelity, evaluating
embodied world models must also reflect their practical value for real-world
applications. We therefore follow the evaluation protocol established by
WorldArena~\cite{shang2026worldarena}, evaluating visual quality, motion naturalness,
and temporal and semantic consistency.

\noindent\textbf{Benchmark construction.}
We construct an Embodied Video Generation benchmark from our generalized
embodied-data mixture, with 200 in-distribution and 100 out-of-distribution tasks.
The benchmark covers diverse objects, scenes, viewpoints, embodiments, action verbs,
and recovery behaviors. The OOD split targets linguistic and compositional
generalization through novel object--verb pairings, paraphrased instructions,
unseen scene configurations, and composed tasks. 80\% of samples include synchronized
control trajectories for action-conditioned evaluation, while the remaining 20\%
follow the text-and-image-to-video setting. This protocol measures whether a model
can generate visually plausible and physically coherent futures from a shared
observation--language context and, when control signals are provided, faithfully
execute the prescribed control sequence.

\begin{figure*}[t]
    \centering
    \includegraphics[width=\textwidth]{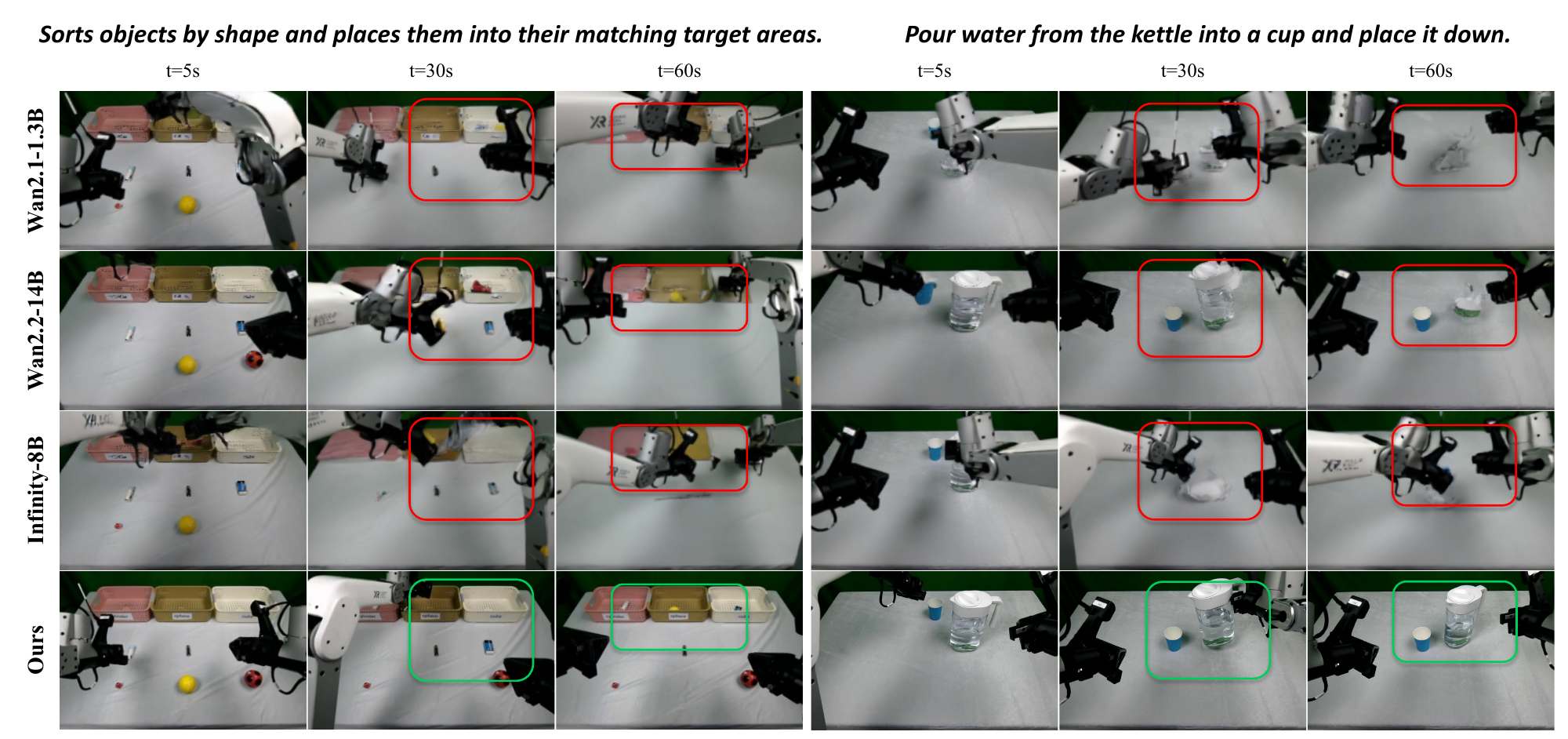}
    \caption{\textbf{Qualitative comparison of long-horizon video rollouts.} Rows correspond to Wan2.1-1.3B, Wan2.2-14B, Infinity-8B, and our model, while columns show snapshots at 5, 30, and 60 seconds. The left task requires sorting tabletop objects into the corresponding baskets, and the right task requires pouring water into a cup. The comparison highlights each model’s ability to preserve scene geometry, object identity, and coherent task progression over extended horizons.}
    \label{fig:embodied-video-generation-qualitative}
\end{figure*}

\noindent\textbf{Main results.}
As shown in Tab.~\ref{tab:embodied-video-generation}, \ours{} consistently improves
upon its InfinityStar backbone and achieves the strongest overall performance on
embodied-relevant dimensions, including motion quality, temporal and semantic
consistency, and physical and conditional fidelity. These gains suggest that
embodied training, together with our scale-aligned action conditioning and
long-horizon training recipe, turns the inherited next-scale video prior into a
stronger physical prior that better preserves coherent motion, geometry, and
robot--object interaction dynamics during generation. Beyond competitive perceptual
quality, \ours{} leads on the core physically grounded and interaction-oriented
criteria, indicating tighter correspondence between prescribed controls and the
resulting manipulation dynamics. Qualitatively, Fig.~\ref{fig:embodied-video-generation-qualitative}
shows that generic video models are more susceptible to motion and interaction drift,
whereas \ours{} maintains more coherent and goal-consistent behavior over long
horizons.

\begin{table*}[t]
    \centering
    \scriptsize
    \setlength{\tabcolsep}{2.2pt}
    \renewcommand{\arraystretch}{1.12}
    \resizebox{\textwidth}{!}{%
    \begin{tabular}{l|cc|ccc|ccc|cccccc}
        \toprule
        \multirow{2}{*}{Model}
        & \multicolumn{2}{c|}{Visual Quality}
        & \multicolumn{3}{c|}{Motion Quality}
        & \multicolumn{3}{c|}{Temporal \& Semantic Consistency}
        & \multicolumn{6}{c}{Physical \& Conditional Fidelity} \\
        \cmidrule(lr){2-3}
        \cmidrule(lr){4-6}
        \cmidrule(lr){7-9}
        \cmidrule(lr){10-15}
        & \makecell{Image\\Quality\\$\uparrow$}
        & \makecell{Aesthetic\\Quality\\$\uparrow$}
        & \makecell{Dynamic\\Degree\\$\uparrow$}
        & \makecell{Flow\\Score\\$\uparrow$}
        & \makecell{Motion\\Smooth.\\$\uparrow$}
        & \makecell{Subject\\Consist.\\$\uparrow$}
        & \makecell{Background\\Consist.\\$\uparrow$}
        & \makecell{Semantic\\Align.\\$\uparrow$}
        & \makecell{Interaction\\Quality\\$\uparrow$}
        & \makecell{Perspectivity\\$\uparrow$}
        & \makecell{Instruction\\Following\\$\uparrow$}
        & \makecell{Trajectory\\Acc.\\$\uparrow$}
        & \makecell{Depth\\Acc.\\$\uparrow$}
        & \makecell{Action\\Following\\$\uparrow$} \\
        \midrule
        InfinityStar~\cite{liu2025infinitystar}
        & 0.602 & 0.321
        & 0.253 & 0.044 & 0.601
        & 0.803 & 0.879 & 0.860
        & 0.484 & 0.768 & 0.406 & \underline{0.251} & 0.749 & -- \\
        Wan2.1-1.3B~\cite{wan2025wan}
        & 0.679 & 0.345
        & 0.329 & 0.098 & 0.672
        & 0.608 & 0.650 & 0.856
        & 0.452 & 0.858 & 0.380 & 0.197 & 0.715 & -- \\
        Wan2.2-14B~\cite{wan2025wan22}
        & 0.690 & 0.337
        & \underline{0.427} & 0.181 & \textbf{0.793}
        & 0.803 & 0.862 & 0.836
        & 0.476 & 0.836 & 0.394 & 0.159 & 0.657 & -- \\
        CogVideoX-5B~\cite{yang2025cogvideox}
        & \textbf{0.718} & 0.429
        & 0.308 & 0.143 & 0.686
        & 0.776 & 0.823 & 0.890
        & 0.400 & \textbf{0.900} & 0.380 & 0.177 & 0.774 & -- \\
        Cosmos3-Nano~\cite{nvidia2026cosmos3}
        & 0.693 & \underline{0.439}
        & 0.425 & \underline{0.208} & 0.701
        & \underline{0.861} & \underline{0.897} & \underline{0.903}
        & \underline{0.516} & 0.874 & \underline{0.410} & 0.202 & \underline{0.816} & \underline{0.044} \\
        \midrule
        \textbf{\ours{}}
        & \underline{0.697} & \textbf{0.453}
        & \textbf{0.435} & \textbf{0.213} & \underline{0.715}
        & \textbf{0.873} & \textbf{0.903} & \textbf{0.912}
        & \textbf{0.546} & \underline{0.885} & \textbf{0.471} & \textbf{0.539} & \textbf{0.831} & \textbf{0.290} \\
        \bottomrule
    \end{tabular}%
    }
    \caption{\textbf{Quantitative evaluation of embodied video generation.}
    Quantitative comparisons against representative foundation video generators.
    \ours{} achieves the strongest interaction quality, instruction following,
    and trajectory accuracy while maintaining competitive visual and semantic
    quality. Bold and underlined values denote the best and second-best results.
    }
    \label{tab:embodied-video-generation}
\end{table*}

\noindent\textbf{Analysis.}
Compared with the generic video baselines and the inherited InfinityStar
backbone, \ours{} performs consistently better on the generalized
embodied benchmark. The improvements in motion dynamics, semantic consistency,
Interaction Quality, Trajectory Accuracy, and Depth Accuracy suggest that
large-scale embodied training, together with our action-grounded and
long-horizon training recipe, adapts the inherited next-scale video prior into
a stronger physical prior over robot motion, object interaction, and contact
evolution. This effect is supported by the diversity of our embodied data
mixture, which spans varied tasks, recovery behaviors, objects, robot
embodiments, camera viewpoints, and language instructions, exposing the model
to a broader distribution of manipulation dynamics.


The results also suggest stronger generalization and action following ability on the compositional OOD
split, which contains novel object--verb pairings, paraphrased instructions,
unseen scene configurations, and new task compositions. Training on diverse
task- and subtask-level language descriptions ties the instruction to coherent
manipulation segments and their visual--action transitions. The resulting
Semantic Alignment and Instruction Following scores suggest that \ours{} can
follow paraphrased and compositional instructions more reliably, rather than
only matching familiar canonical descriptions.

\begin{figure}[t]
    \centering
    \captionsetup{width=\textwidth}
    \includegraphics[width=\textwidth]{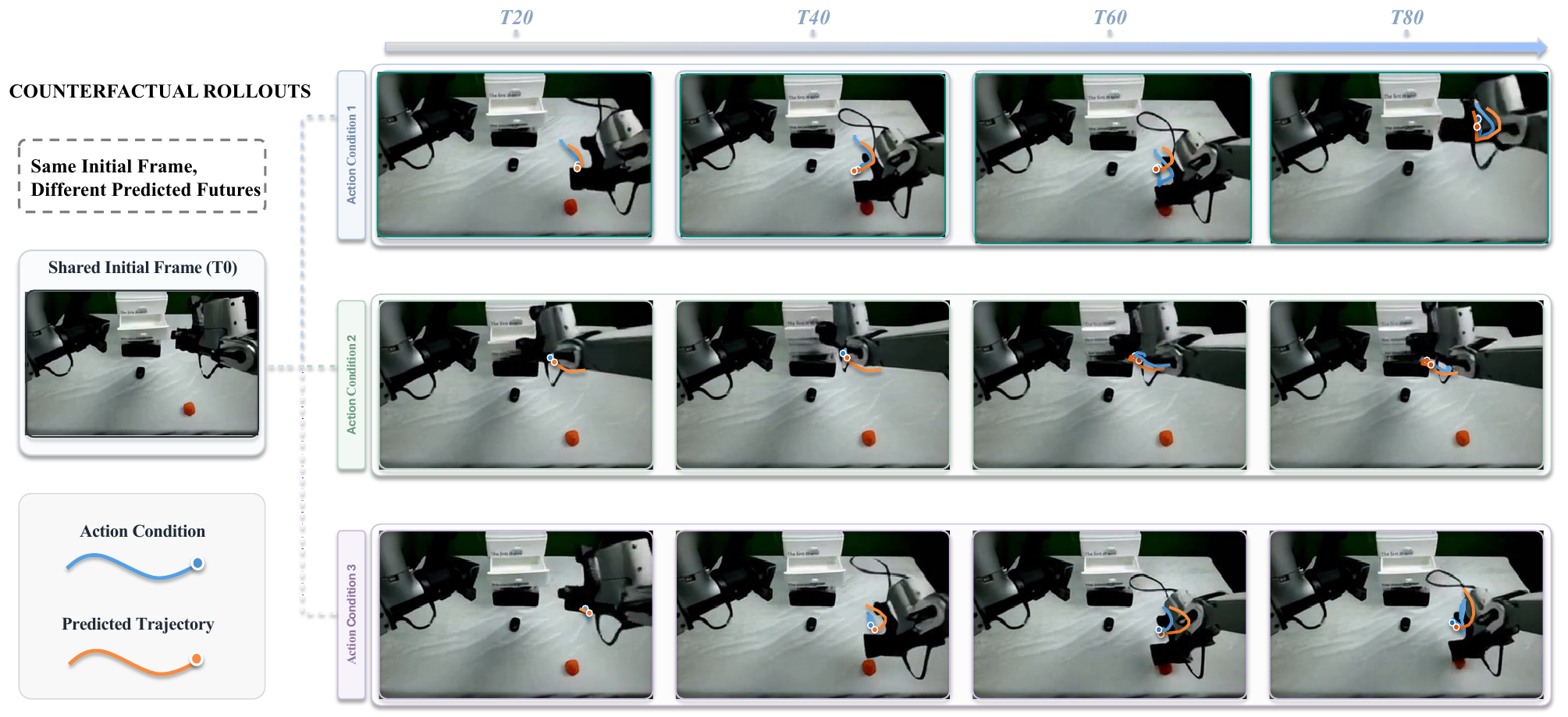}
    \caption{\textbf{Counterfactual action-conditioned rollouts.}
    Starting from the same initial observation, we vary only the commanded
    action and autoregressively generate the corresponding future. Blue and orange curves denote commanded and realized trajectories, respectively.}
    \label{fig:counterfactual-action-rollouts}
\end{figure}

\subsection{Action-Conditioned Video Generation Evaluation}
\label{sec:action-condition-eval}

\noindent\textbf{Evaluation protocol.}
To evaluate the model's action fidelity and action-following capability, we conduct a dedicated action-conditioned generation evaluation.
During action-conditioned evaluation, the model is additionally provided with a synchronized control trajectory specifying the intended motion over varying spatial ranges and throughout the prediction horizon.
We follow WorldArena and report both \emph{Action Following} and \emph{Trajectory Accuracy}.
Action Following evaluates whether distinct action conditions induce distinguishable future outcomes from the same visual context, thereby measuring the model's sensitivity to the prescribed control branch.
Trajectory Accuracy instead measures the spatiotemporal agreement between the robot motion realized in the generated video and the corresponding reference trajectory.
Together, the two metrics assess whether the model not only responds to action inputs, but also realizes them through geometrically and temporally consistent motion.



\noindent\textbf{Analysis.}
As shown in Tab.~\ref{tab:embodied-video-generation}, \ours{} performs favorably on both Action Following and Trajectory Accuracy.
Its strong Action Following score indicates that changing the action condition produces meaningfully different visual futures, rather than minor appearance variations around a single scene-driven continuation.
At the same time, the improved Trajectory Accuracy shows that these counterfactual differences are aligned with the intended robot motion: the generated end-effector follows the prescribed direction and temporal evolution more closely.
The two results together suggest that \ours{} both selects the future specified by the control input and faithfully renders its visual consequences.

We evaluate whether the model can generate action-faithful video sequences under different action conditions while keeping the initial visual observation fixed.
The visual-to-action attention is primarily concentrated around the active gripper, the manipulated object, and nearby interaction regions, and shifts consistently as the manipulation progresses.
This spatially adaptive behavior is consistent with our scale-aligned conditioning mechanism, which routes action information to the visual tokens responsible for local motion and contact dynamics.
Together, these results demonstrate that the model is sensitive to the prescribed control inputs and can generate futures with strong action-following fidelity and accurate trajectory realization.

\FloatBarrier
\noindent\textbf{Action-conditioned Counterfactual Reasoning.}
We further test whether the visual dynamics learned by \ours{} are genuinely
conditioned on the controls, rather than a single continuation determined
primarily by the initial scene and task prior. We hold the initial observation
and language instruction fixed and intervene only on the future control
trajectory, constructing counterfactual action branches.
As shown in Fig.~\ref{fig:counterfactual-action-rollouts}, each row starts
from the same visual state but receives a different commanded trajectory. The
blue dashed curves denote the prescribed end-effector paths, while the orange
curves show the realized trajectories recovered from the generated rollouts.

This capability is further supported by the intervention-rich composition of our training data. Beyond successful demonstrations, the mixture contains extensive human takeovers, recovery
behaviors, and failed executions, exposing the model to similar visual states
followed by different controls and outcomes. Takeover and recovery
segments provide alternative and corrective continuations, while failure
trajectories prevent the model from collapsing toward an unrealistically
optimistic successful future. Together with scale-aligned action conditioning,
these examples encourage \ours{} to distinguish and faithfully realize
multiple action-dependent futures.

\subsection{Long-Horizon Streaming Rollout Evaluation}
\label{sec:long-horizon-memory-eval}

\noindent\textbf{Evaluation protocol.}
Short-clip quality alone does not determine whether a world model remains
reliable once its own predictions are recursively committed as future context.
We therefore evaluate long-horizon action-conditioned generation under a
temporally segmented replay protocol, with rollouts extending to 60 seconds.
Each model starts from the same synchronized multi-view observation and
receives the same prescribed control sequence; after initialization, every
visual clip is generated autoregressively and committed through the model's
streaming state transition, and ground-truth future frames are never fed
back---they serve only as references for evaluation.

\begin{figure}[t]
    \centering
    \captionsetup{width=\textwidth}
    \includegraphics[width=\textwidth]{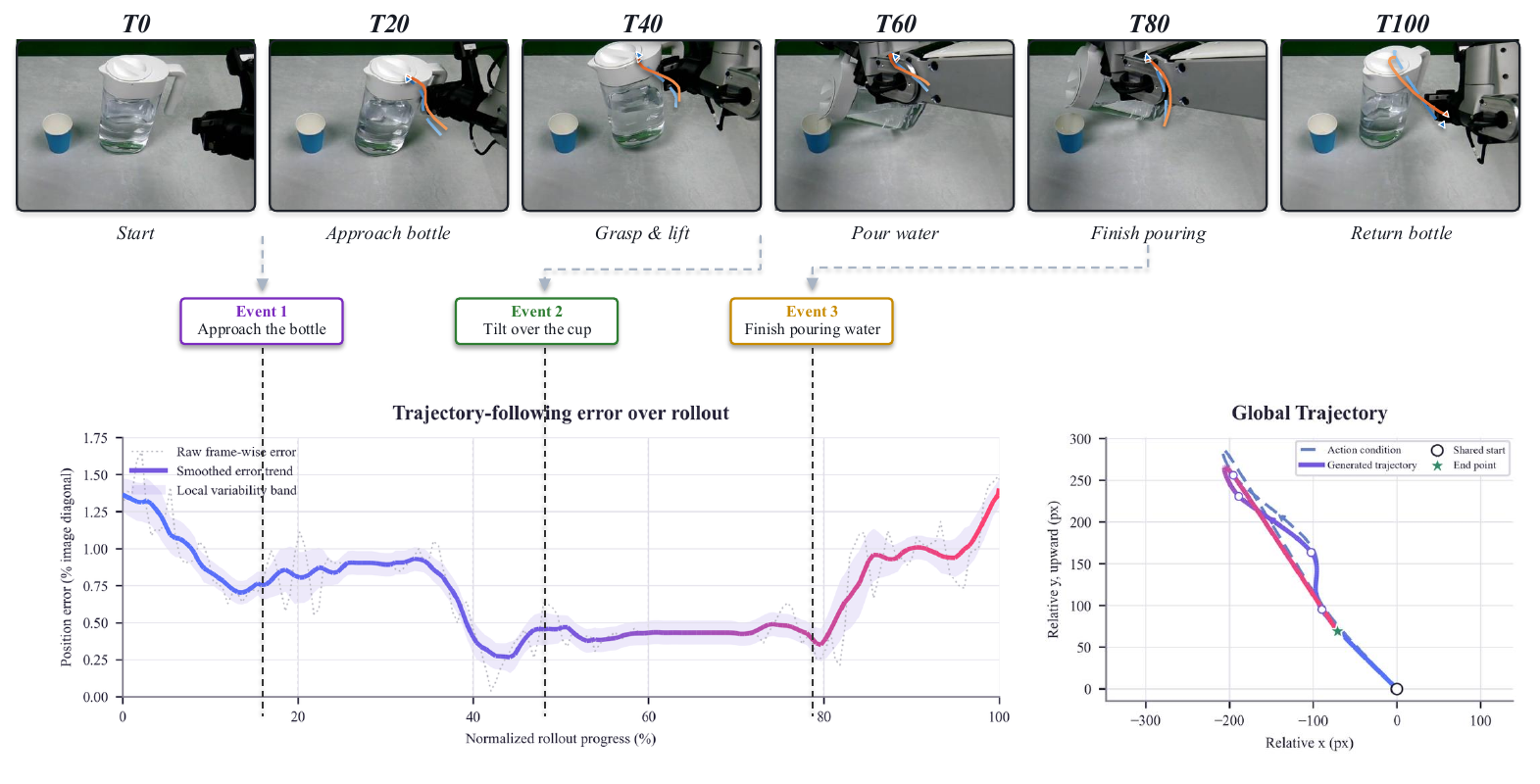}
    \caption{\textbf{Long-horizon action following in a streaming rollout.}
    For a single continuous rollout, shown as keyframes with the
    overlaid end-effector path, the frame-wise trajectory-following error
    remains below $0.5\%$ of the image diagonal throughout the contact-rich
    pouring phase, while the generated global path follows the commanded
    condition.}
    \label{fig:long-horizon-action-following}
\end{figure}

\noindent\textbf{Streaming rollout visualization.}
Fig.~\ref{fig:long-horizon-action-following} visualizes one continuous
rollout at representative control steps. The commanded path is propagated
through successive generated clips, while the realized end-effector trajectory
and interaction state are measured from the rollout itself. Compared with the
recent-clip-only baseline, the time--scale streaming model accumulates
substantially less trajectory error and remains close to the prescribed global
path across the full sequence.

We compare the complete model against three controlled memory variants.
\emph{Without dream forcing} retains the same inference-time memory mechanism
but removes training on self-generated contexts, isolating robustness to
model-induced history. \emph{Recent-clip only} discards distant records and
conditions only on the latest generated state, isolating the contribution of
long-range state retention. \emph{Full-history KV} preserves all historical
visual states at uniform precision, a quality-oriented reference whose memory
and attention cost grow with rollout duration. Together, these variants
separate the effects of self-context training, time--scale retention, and
unrestricted historical context.

\begin{figure*}[t]
    \centering
    \includegraphics[width=\textwidth]
    {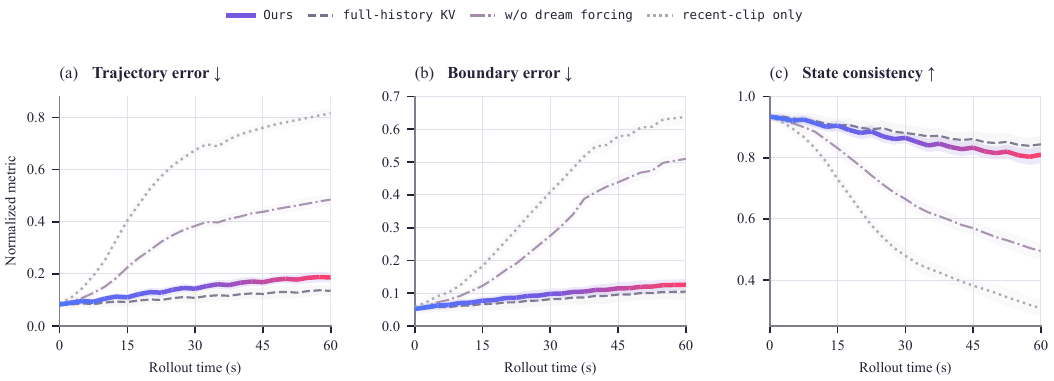}
    \caption{\textbf{Long-horizon action-conditioned rollout dynamics.}
    All variants start from the same synchronized observation and follow the
    same prescribed control sequence. (a) End-effector trajectory error
    measures action-grounded motion relative to the matched reference.
    (b) Cross-clip boundary error measures discontinuity when a generated clip
    is committed as context for the next one. (c) Persistent-state consistency
    measures retention of robot, object, and contact state. Curves are
    aggregated over successive rollout chunks.}
    \label{fig:long-horizon-rollout-dynamics}
\end{figure*}

\noindent\textbf{Action, state, and boundary retention.}
Fig.~\ref{fig:long-horizon-rollout-dynamics} separates three failure modes
that a single perceptual score would conflate. We obtain trajectory error by
tracking the generated end effectors and aligning their paths with the
reference control trajectory using normalized dynamic time warping. Persistent
state consistency evaluates the robot pose, manipulated object, contact
relation, and task-relevant object state within each rollout window.
Cross-clip boundary error is computed around each commit boundary from
robot- and object-centric motion tracks, exposing visible jumps that may be
hidden by an average full-video metric. Together, the panels test whether
action effects remain correct, whether their resulting state is remembered,
and whether recursive generation advances continuously.

\begin{figure*}[t]
    \centering
    \includegraphics[width=\textwidth]
    {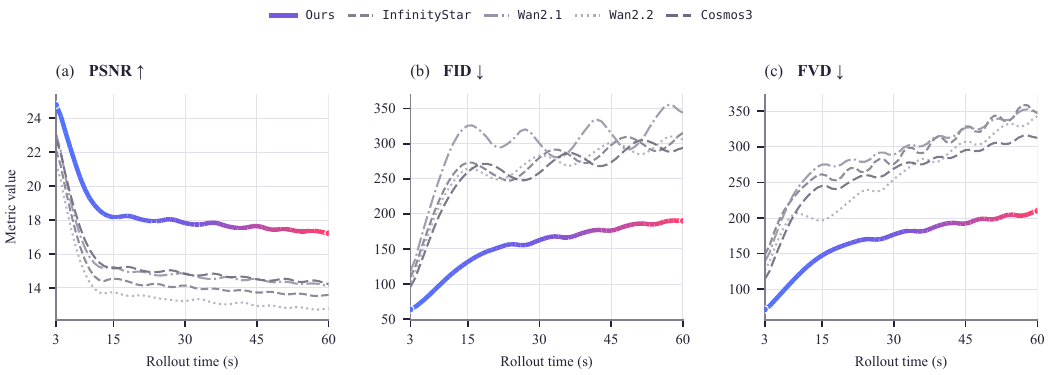}
    \caption{\textbf{Perceptual quality over a 60-second autoregressive
    rollout.}
    PSNR ($\uparrow$), FID ($\downarrow$), and FVD ($\downarrow$) are measured
    in successive rollout windows. Across all windows, \ours{} achieves higher
    PSNR and lower FID and FVD than others.}
    \label{fig:long-horizon-perceptual-quality}
\end{figure*}

\noindent\textbf{Perceptual degradation over rollout time.}
We report quality as a function of elapsed rollout time so that late-stage
identity collapse or texture accumulation cannot be hidden by strong early
chunks. PSNR is computed from temporally aligned generated--reference pairs,
whereas FID and FVD are computed in successive three-second windows on a
common held-out set. The plotted curves use a three-window moving average,
while all aggregate comparisons use the underlying window scores.
Fig.~\ref{fig:long-horizon-perceptual-quality} therefore complements the
physical diagnostics in
Fig.~\ref{fig:long-horizon-rollout-dynamics}: the former measures how visual
fidelity changes, while the latter determines whether apparently plausible
motion remains action- and state-consistent.

\begin{wrapfigure}[18]{r}{0.46\linewidth}
    \vspace{-10pt}
    \centering
    \includegraphics[width=\linewidth]
    {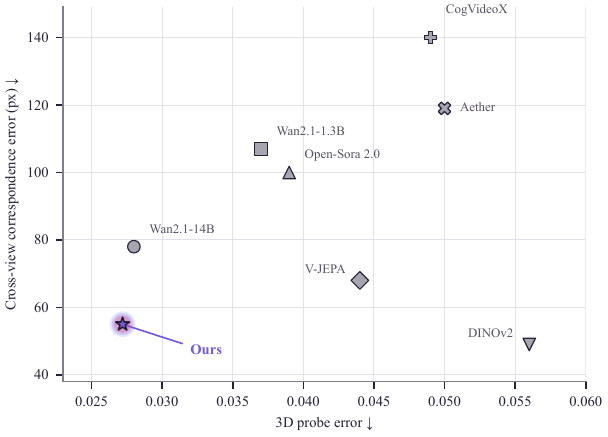}
    \vspace{-6pt}
    \caption{\textbf{Cross-view consistency.}
    \ours{} achieves low 3D-probe and correspondence errors (both
    $\downarrow$), indicating 3D-aware and view-consistent representations.}
    \label{fig:long-horizon-multiview-consistency}
    \vspace{-10pt}
\end{wrapfigure}

\noindent\textbf{Cross-view consistency.}
Long-horizon stability must hold for one physical state rather than for each
camera independently. We therefore evaluate synchronized triplets from one
head camera and two wrist cameras. A frozen linear probe measures the
recoverability of 3D structure from generated representations, while calibrated
point matches measure cross-view correspondence error in pixels. Each point in
Fig.~\ref{fig:long-horizon-multiview-consistency} aggregates valid
synchronized windows over the evaluated rollouts. The favorable lower-left
region requires both decodable spatial structure and agreement across
viewpoints, ruling out a failure mode in which each camera remains individually
plausible while the generated robot or object state becomes mutually
inconsistent.

Taken together, we evaluate complementary aspects of
the same streaming process. The rollout diagnostics isolate action, persistent
state, and boundary errors; the temporal perceptual metrics expose cumulative
visual degradation; and the multi-view probe measures whether synchronized
outputs remain spatially compatible across cameras.

\FloatBarrier
\subsection{Reinforcement-Learning Evaluation}
\label{sec:rl-eval}

We isolate this stage with a controlled ablation: the supervised checkpoint is
compared with its reward-aligned counterpart under identical held-out initial
observations, action conditions, and sampling parameters, and both models
generate autoregressively from their own visual history---the same on-policy
context in which the rewards were optimized. All metrics reuse the protocols
already defined in this section: action following and trajectory accuracy
from Sec.~\ref{sec:action-condition-eval}, and persistent-state
consistency and cross-clip boundary error over recursive rollouts from
Sec.~\ref{sec:long-horizon-memory-eval}.

The ablation shows consistent gains that concentrate exactly on the two axes
the frozen reward scorers target. On the action side, alignment improves
action following from $0.264$ to $0.290$ and trajectory accuracy from $0.512$
to $0.539$; qualitatively, under the same commanded end-effector trajectory
the supervised checkpoint tends to accumulate directional error late in the
rollout, whereas the aligned model keeps the realized motion close to the
prescribed path throughout. On the long-horizon side, cross-clip boundary
error drops from $0.118$ to $0.104$, and persistent-state consistency
improves in tandem, indicating that the long-horizon fidelity reward
mitigates the slow drift and commit-boundary discontinuities that supervised
training alone leaves behind. The magnitude of these gains is bounded by
design: reward optimization starts from a strong supervised checkpoint and is
explicitly constrained to stay close to it, so its role is to correct
residual on-policy failure modes rather than to relearn the dynamics.

Two properties of the ablation matter beyond the raw numbers. First,
appearance metrics such as subject consistency and image quality remain
unchanged within evaluation noise, indicating that alignment reallocates
probability mass among dynamics-consistent visual futures rather than
trading appearance for reward; we attribute this stability to the KL
constraint and interleaved real-trajectory replay of
Sec.~\ref{sec:fidelity-visual-alignment}. Second, because the rewards
deliberately exclude any notion of task progress or success, the gains
reported here cannot be an artifact of optimizing toward the evaluation
target; the closed-loop evaluation of
Sec.~\ref{sec:closed-loop-policy-consistency}, which uses the
reward-aligned checkpoint throughout, therefore serves as an independent
check that these dynamics-level gains carry over to policy-evaluation
utility.

The same alignment stage also benefits the action expert on the physical robot,
which is the most demanding transfer we test. Reward optimization updates only
the visual generator and its conditioning adapters, and the action expert
receives no reward gradient of its own; it consumes the aligned dynamics solely
because it predicts controls from the same committed causal state
(Sec.~\ref{sec:world-action-policy}). Running the identical action-expert
co-training on the supervised and on the reward-aligned generator, and
deploying both under the protocol of
Sec.~\ref{sec:real-robot-evaluation}, raises average real-robot Task
Progress from $64.6$ to $69.1$. The gain concentrates where the two scorers
apply: on contact-rich transfer segments, which the action-following reward
pushes toward the commanded motion, and on multi-stage tasks, whose state must
survive successive commit boundaries under the long-horizon reward. Since
neither reward observes actions, task progress, or task success, this is
evidence that alignment sharpens a physical prior---how a scene must move to be
consistent with a given control---and that this prior is exactly what the
action expert reads when it selects controls. Reward-based alignment of visual
dynamics therefore improves the world model as a simulator and as a policy at
the same time, without a control-specific reward.

\begin{figure*}[t]
    \centering
    \includegraphics[width=\textwidth]{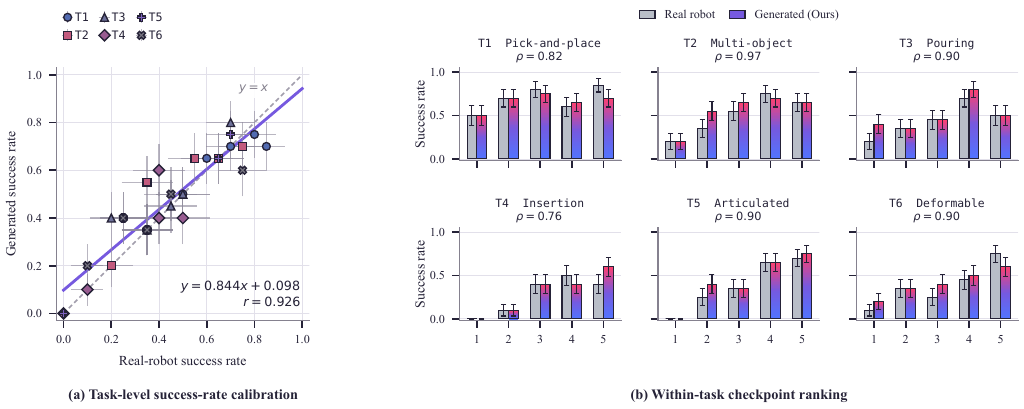}
    \caption{\textbf{Closed-loop calibration and within-task checkpoint ranking.}
    (a) Task-level success-rate calibration between matched real-robot and
    generated rollouts. Each marker represents one task--checkpoint pair;
    the dashed diagonal denotes perfect agreement and the solid line denotes
    the least-squares fit.
    (b) Within-task ranking across five policy checkpoints. Bars compare real-robot and generated success rates, with error
    bars denoting per-cell binomial standard errors. The reported $\rho$
    values are per-task Spearman rank correlations.}
    \label{fig:closed-loop-calibration-ranking}
\end{figure*}

\FloatBarrier
\subsection{Closed-Loop Policy Consistency}
\label{sec:closed-loop-policy-consistency}
Replay fidelity cannot determine whether a world model can evaluate robot
policies, because rollout errors may alter subsequent actions and outcomes.
We therefore execute the same frozen policies in \ours{} and on the physical
robot, and compare their closed-loop outcomes and checkpoint rankings.

\begin{table}[h]
    \centering
    \footnotesize
    \setlength{\tabcolsep}{2pt}
    \renewcommand{\arraystretch}{1.0}
    \begin{tabular}{@{}l p{0.30\columnwidth} p{0.46\columnwidth}@{}}
        \toprule
        Task & Task type & Subtasks \\
        \midrule
        T1 & Put glasses on shelf &
        Grasp glasses; Place glasses upright on shelf \\
        T2 & Put fruits in basket &
        Grasp fruit in order; Place fruit in basket in order \\
        T3 & Pour water into cup &
        Grasp bottle; Align with cup;
        Pour water; Set bottle down \\
        T4 & Put ring on rod &
        Grasp ring; Align ring with rod; Slide ring onto rod \\
        T5 & Put item in drawer &
        Open drawer; Grasp item; Place item in drawer; Close drawer \\
        T6 & Put garment in hamper &
        Grasp garment; Move over hamper; Place garment in hamper \\
        \bottomrule
    \end{tabular}
    \caption{\textbf{Tasks for closed-loop policy consistency evaluation.}
    A variety of manipulation tasks are decomposed into ordered subtasks to assess
    execution consistency beyond final task success.}
    \label{tab:closed-loop-policy-tasks}
\end{table}
\noindent\textbf{Evaluation protocol.}
Five WALL-WM checkpoints spanning low to high validation competence are
evaluated on six held-out tasks with 20 matched initial configurations,
yielding $6\times5\times20=600$ generated--real rollout pairs
(Tab.~\ref{tab:closed-loop-policy-tasks}). Generated rollouts follow
Sec.~\ref{sec:training-inference} without ground-truth feedback after
initialization. Blinded annotators label outcomes and ordered-subtask
completion; unrecoverable generation artifacts count as failures. We report
task-level calibration, within-task ranking, and episode-level agreement with
stratified-bootstrap 95\% confidence intervals.

\begin{figure*}[t]
    \centering
    \includegraphics[width=0.98\textwidth]{
        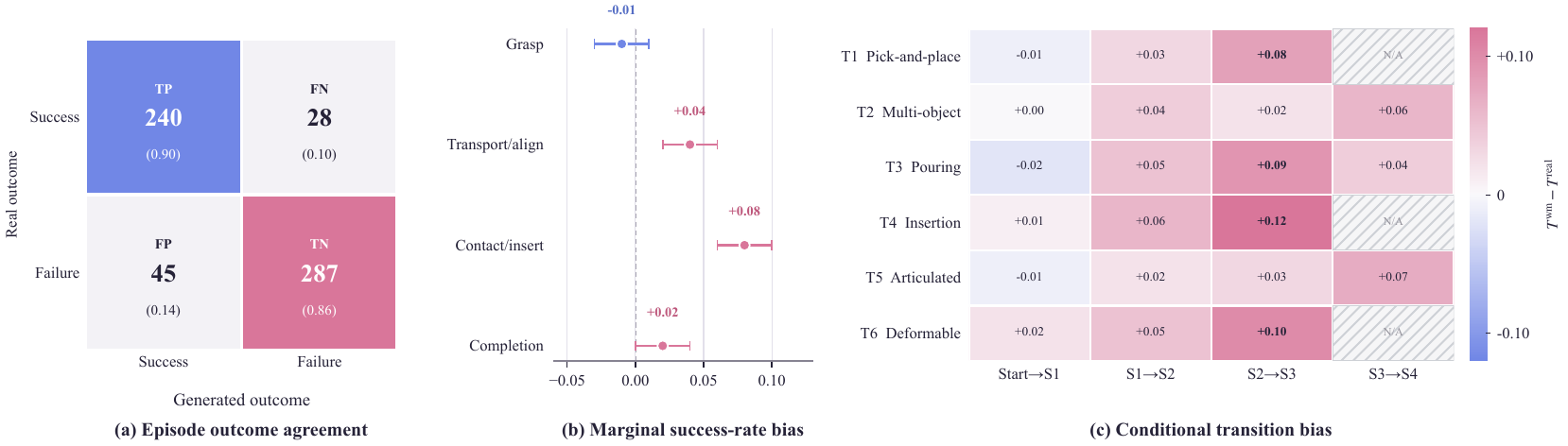
    }
    \caption{\textbf{Closed-loop fidelity across outcomes and subtask
    transitions.}
    (a) Confusion matrix of \ours{} and real-robot outcomes for 600 paired
    rollouts; entries give counts and row-normalized rates.
    (b) Marginal generated-minus-real success-rate bias at each execution
    stage; error bars are binomial standard errors.
    (c) Conditional transition differences
    $T^{\mathrm{wm}}-T^{\mathrm{real}}$ identify where discrepancies first
    emerge; hatching indicates unavailable task--transition combinations.
    \ours{} matches the real outcome in 527 episodes, with the remaining
    optimism concentrated at contact/insertion stages.}
    \label{fig:closed-loop-agreement-bias}
\end{figure*}

\begin{table*}[t]
    \centering
    \small
    \setlength{\tabcolsep}{2.2pt}
    \renewcommand{\arraystretch}{1.1}
    \begin{tabular}{l cccc c ccc c cccc}
        \toprule
        & \multicolumn{4}{c}{Calibration (30 cells)}
        && \multicolumn{3}{c}{Within-task ranking}
        && \multicolumn{4}{c}{Episode agreement (600 pairs)} \\
        \cmidrule{2-5} \cmidrule{7-9} \cmidrule{11-14}
        Model
        & MAE $\downarrow$ & Bias $\rightarrow 0$ & Slope $\rightarrow 1$
        & $r$ $\uparrow$
        && Pair.\ Acc $\uparrow$ & Regret $\downarrow$ & $\bar\rho$ $\uparrow$
        && Bal.\ Acc $\uparrow$ & Succ.\ Rec.\ $\uparrow$
        & Fail.\ Rec.\ $\uparrow$ & FPR $\downarrow$ \\
        \midrule
        \textbf{\ours{}}
        & 0.062 & $+0.028$ & 0.84 & 0.93
        && 0.89 & 0.025 & 0.88
        && 0.88 & 0.90 & 0.86 & 0.14 \\
        \bottomrule
    \end{tabular}
    \caption{\textbf{Closed-loop outcome alignment.}
    Generated and real rollouts are compared at three levels: success-rate
    calibration across 30 task--checkpoint cells, checkpoint ranking over all
    untied within-task pairs, and outcome agreement over 600 matched episodes.
    Ranking metrics include pairwise accuracy, mean selection regret
    (Eq.~\eqref{eq:closed-loop-regret}), and mean per-task Spearman
    $\bar\rho$.}
    \label{tab:closed-loop-policy-consistency}
\end{table*}

\noindent\textbf{Results and analysis.}
\textit{Task-level calibration.}
Across 30 task--checkpoint cells, generated and real success rates have an MAE
of $0.062$ (95\% CI $[0.06,0.11]$) and signed bias of $+0.028$
$[-0.01,0.06]$. The fitted calibration line has slope $0.84$
$[0.65,0.97]$
(Fig.~\ref{fig:closed-loop-calibration-ranking}(a)), indicating mild
overestimation of weaker checkpoints. With only $N=20$ trials per cell, we
interpret aggregate trends rather than individual cell differences.

\textit{Within-task ranking.}
To avoid confounding policy quality with task difficulty, checkpoints are
ranked only within each task. \ours{} preserves $89\%$ $[0.76,0.95]$ of the
59 untied real-robot pairwise orderings, attains mean per-task Spearman
$\bar\rho=0.88$, and selects the best checkpoint on four of six tasks
(Fig.~\ref{fig:closed-loop-calibration-ranking}(b)). Regret is
\begin{equation}
    R_q=\max_p S^{\mathrm{real}}_{q,p}
    -S^{\mathrm{real}}_{q,\hat p_q},
    \quad \hat p_q=\operatorname*{arg\,max}_p S^{\mathrm{wm}}_{q,p},
    \label{eq:closed-loop-regret}
\end{equation}
with mean $0.025$ and worst-task value $0.10$; both selection errors remain
within per-cell sampling uncertainty.

\textit{Episode-level agreement.}
Treating each generated outcome as a binary prediction of its matched real
outcome (Fig.~\ref{fig:closed-loop-agreement-bias}(a)), balanced accuracy
over the 600 pairs is $0.88$ $[0.83, 0.93]$ (MCC $0.76$), with success
recall $0.90$ and failure recall $0.86$. The false-positive rate
$P(\text{generated success}\mid\text{real failure})$---the dominant risk
mode of an optimistic evaluator---is $0.136$ $[0.09, 0.19]$: in $45$ of the
$332$ real failures, the generated rollout still ends in success, the error
direction that silently promotes weak checkpoints.

\textit{Subtask-level bias.}
The residual bias is stage dependent
(Fig.~\ref{fig:closed-loop-agreement-bias}(b,c)): grasp is nearly calibrated
($-0.01$), transport and alignment are mildly optimistic ($+0.04$), and
contact dominates the discrepancy ($+0.08$) before it falls at completion
($+0.02$). Conditional transition bias peaks at $+0.12$ for insertion,
showing that the model most often overestimates progress through
contact-intensive transitions.

\FloatBarrier
\subsection{Real-Robot Evaluation}
\label{sec:real-robot-evaluation}

\noindent\textbf{Experimental setup.}
We evaluate whether the predictive representations learned by \ours{} transfer to
closed-loop physical control. We deploy the co-trained action expert of Sec.~\ref{sec:world-action-policy} 
on our tabletop bimanual platform with synchronized
head and wrist cameras. At each chunk boundary, the policy takes the latest
multi-view observation, task instruction, and proprioception, and predicts the
next executable action chunk. This complements
Sec.~\ref{sec:closed-loop-policy-consistency}: there, a frozen external policy probes
the fidelity of model-generated dynamics, whereas here the action branch of
\ours{} directly controls the physical robot.

We evaluate several tabletop tasks covering direct manipulation
(\emph{Arrange Cup Inverted Triangle}, \emph{Put Spoon to Bowl},
\emph{Put Ring onto Rod}, and \emph{Pick Items into Basket}) and
instruction-conditioned manipulation (\emph{Sort Headphone},
\emph{Classify Items as Shape}, and \emph{Press Button in Order}).
We report \emph{Task Progress}, a dense $0$--$100$ rubric score that credits
meaningful intermediate completion, together with strict success rate.
We compare against $\pi_{0.5}$~\cite{black2025pi05},
LingBot-VA~\cite{li2026causal}, and DreamZero~\cite{ye2026world} under identical
task definitions, observations, scene randomization, and scoring rubrics,
without privileged state or scoring feedback.

\begin{figure*}[t]
\centering
\includegraphics[width=\textwidth]
{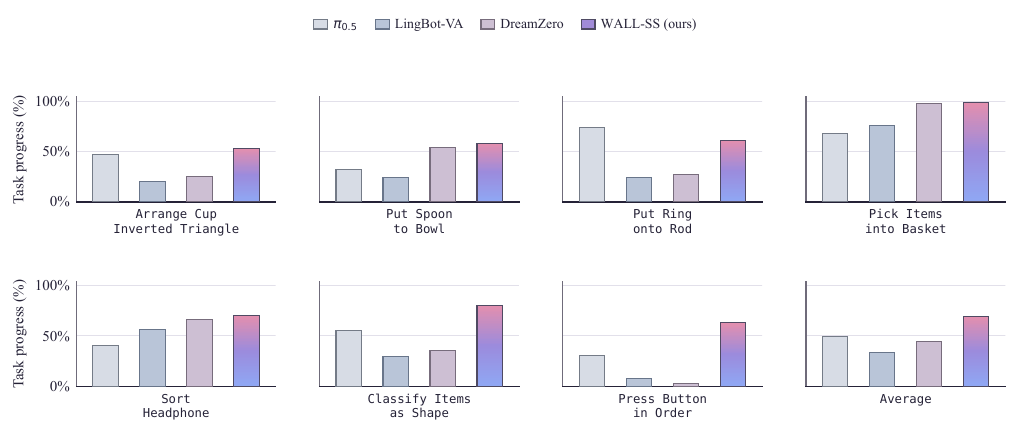}
\caption{\textbf{Diverse real-robot manipulation.}
Task Progress on tabletop tasks covering direct physical manipulation,
pick-and-place, insertion, instruction-conditioned sorting and classification,
and sequential interaction. The final group reports the unweighted task
average.}
\label{fig:real-robot-diverse-manipulation}
\end{figure*}

\noindent\textbf{Results.}
As shown in Fig.~\ref{fig:real-robot-diverse-manipulation}, \ours{} achieves an
average Task Progress of $69.1$, outperforming $\pi_{0.5}$ ($49.6$),
DreamZero ($44.1$), and LingBot-VA ($34.0$), and ranks first on these tasks. A key distinction is that action prediction and future generation share
the same committed causal state, allowing the action objective to directly shape
representations used for predictive world modeling.

The gains are pronounced on direct-manipulation tasks: \ours{} reaches $98.5$
on \emph{Pick Items into Basket}, and obtains the best scores on
\emph{Arrange Cup Inverted Triangle} ($53$) and \emph{Put Spoon to Bowl} ($58$).
These tasks require complete approach--grasp--transfer--release transitions,
consistent with the tighter action--future coupling.
The advantage is also clear when task history matters. \ours{} obtains $70$ on
\emph{Sort Headphone}, $80$ on \emph{Classify Items as Shape}, and $63$ on
\emph{Press Button in Order}, compared with best-baseline scores of $66$, $55$,
and $31$, respectively. In particular, ordered execution requires tracking
previously completed subgoals; our bounded time--scale memory preserves recent interactions and coarser
long-term state together with their associated actions.
Overall, these results show that the predictive state learned by our next-scale
autoregressive world model is not only useful for future generation, but also
provides action-relevant representations for long-horizon physical control.

\section{Discussion}
\label{sec:discussion}

\paragraph{From visual prediction to interactive understanding.}
A world model becomes useful for robotics when its predictions form a coherent
account of how observations, actions, and outcomes relate over time. This
shifts the objective beyond synthesizing plausible videos toward representing
alternative futures, preserving the consequences of interaction, and
supporting decisions under changing conditions. Progress should therefore be
judged by the breadth of interactions a model can explain and the consistency
with which its internal world supports different downstream uses. We view this
transition from appearance generation to reusable interactive understanding as
a central step toward more general embodied systems.

\paragraph{Continual evaluation and improvement.}
Embodied systems operate in distributions that evolve with new environments,
policies, and human expectations. A static benchmark can reveal local
strengths, but cannot fully characterize how a predictive model behaves as
tasks, data, and deployment conditions change. Future research should combine
broader evaluation coverage with uncertainty, feedback, and continual
refinement, so that failures become informative experience rather than
isolated scores. Shared protocols that connect predictions, decisions, and
physical outcomes across platforms will be important for making world models
dependable components of the broader robot-learning ecosystem.

\section*{Contributors}
\label{sec:contributors}

WALL-SS is a collaborative effort of the X Square Robot team. The full contributor list is given below; $^{\ast}$ denotes core contributors, $^{\dagger}$ denotes the project lead, and $^{\ddagger}$ denotes the corresponding author.

\vspace{0.5em}
\noindent
Maeve Zhang$^{\ast}$, Rain Sun$^{\ast}$, Xiang Wang$^{\ast}$, Cyril Zhang$^{\ast}$, Shalfun Li$^{\ast\dagger}$, Meng Cao, Howard Lu, Ethan Chen, Harry jhou, KZ Zheng, Lights Shi, Regis Cheng, Lorenzin, Robert Wang, Victor Yao, Gody Li, Elise Mon, Yohann Tang, Ryan Yu, PS Zhang, Vincent Chen, Hang Su, Roy Gan, Hao Wang$^{\ddagger}$, Qian Wang.

\clearpage
\bibliographystyle{assets/plainnat}
\bibliography{paper}

\clearpage
\appendix

\end{document}